%% file: main.tex
\documentclass[11pt, twoside]{article}
\usepackage{arxiv}
\usepackage{amsthm}
\usepackage[utf8]{inputenc}
\usepackage{amsmath}
\usepackage{array}
\usepackage{graphicx}
\usepackage[hyphens]{url}
\usepackage{multirow}
\usepackage{hyperref}
\usepackage{float}
\usepackage{subfig}
\usepackage{multicol}
\usepackage{mathtools}
\usepackage{schemata}
\usepackage{dsfont}
\usepackage{tabularx}
\usepackage{todonotes}
\usepackage{algorithm} 
\usepackage[inline]{enumitem}
\usepackage{booktabs}
\usepackage{amssymb}
\usepackage{listings}
\usepackage[most]{tcolorbox}
\usepackage{amsthm}
\usepackage[notransparent]{svg}

\usepackage{amsfonts}
\usepackage{comment}
\usepackage{bbm}
\usepackage{placeins}
\usepackage{color,colortbl}
\usepackage{afterpage}
\usepackage{pdflscape}
\usepackage{hyperref}
\usepackage{cleveref}
\usepackage{acronym}
\usepackage[numbers]{natbib}
\usepackage{pifont}
\usepackage[font=footnotesize]{caption}
\usepackage{xcolor}
\usepackage{fancyhdr}
\colorlet{BLACK}{black}
\crefformat{footnote}{#2\footnotemark[#1]#3}

\definecolor{LightGray}{gray}{0.9}

\newtheorem{example}{Example}
\def\eatcell#1\unskip{}
\newcolumntype{E}{>{\eatcell}c@{}}
\newcommand{\chatbox}[1]{%
  \begin{tcolorbox}[colframe=black!70!white,colback=white,arc=4pt,boxrule=0.5pt,fontupper=
  \small,top=2pt,left=2pt,right=2pt,bottom=2pt]
  #1
  \end{tcolorbox}
}

\newcommand{\examplebox}[2]{%
  \begin{tcolorbox}[colframe=black!70!white,colback=blue!2!white,arc=4pt,boxrule=0.5pt,fontupper=
  \small,top=2pt,left=2pt,right=2pt,bottom=2pt,title=#1]
  #2
  \end{tcolorbox}
}

\newcommand{\czm}[1]{\textcolor{black}{#1}}
\newcommand{\new}[1]{\textcolor{black}{#1}} 
\newcommand{\rms}[1]{\textcolor{black}{#1}}
\newcommand{\ph}[1]{\textcolor{black}{#1}}

\title{Large language models for crowd decision making based on prompt design strategies using ChatGPT\ph{: models, analysis and challenges}}
\author{David Herrera-Poyatos \And 
        Cristina Zuheros \And 
        Rosana Montes \And 
        \bigskip
        Francisco Herrera \And \\
D. Herrera-Poyatos, C. Zuheros and F. Herrera are with the \\Department of Computer Sciences and Artificial Intelligence, and the \\Andalusian Research Institute in Data Science and Computational Intelligence (DaSCI)\\ 
 University of Granada, 18071 Granada, Spain.  \\
        \bigskip  \textit{divadhp@ugr.es; czuheros@ugr.es; herrera@decsai.ugr.es}\\
R. Montes is with the Department of Software Engineering, and the\\Andalusian Research Institute in Data Science and Computational Intelligence (DaSCI)\\ University of Granada, 18071, Granada, Spain. \\\textit{rosana@ugr.es}}

\hypersetup{
pdftitle={Large language models for crowd decision making based on prompt design strategies using ChatGPT: models, analysis and challenges},
pdfsubject={LLM for CDM using ChatGPT},
pdfauthor={David Herrera-Poyatos, 
        Cristina Zuheros,,
        Rosana Montes,
        Francisco Herrera},
pdfkeywords={Large Language Models, ChatGPT, Crowd Decision Making, Sentiment Analysis, Prompt Design Strategies},
}

\date{March, 2024}

\renewcommand{\headeright}{}
\renewcommand{\undertitle}{}
\begin{document}
\maketitle

\begin{center}
\textit{This work has been submitted to the IEEE for possible publication. Copyright may be \\transferred without notice, after which this version may no longer be accessible.}
\end{center}


\begin{abstract}
Social Media and Internet have the potential to be exploited as a source of opinion to enrich Decision Making solutions. Crowd Decision Making (CDM) is a methodology able to infer opinions and decisions from plain texts\new{, such as reviews published in social media platforms,} by means of Sentiment Analysis. Currently, the emergence and potential of Large Language Models (LLMs) lead us to explore new scenarios of automatically understand written texts, also known as natural language processing. This paper analyzes the use of ChatGPT based on prompt design strategies to assist in CDM processes to extract opinions and make decisions. We integrate ChatGPT in CDM \czm{processes} as a flexible tool that infer the \czm{opinions} expressed in texts, providing numerical \new{or linguistic evaluations} \ph{where the decision making models are based on the prompt design strategies}. \czm{We include a multi-criteria decision making scenario with a category ontology for criteria.} \czm{We also} consider ChatGPT as an end-to-end CDM \ph{model} able to provide a general opinion \czm{and score} on the alternatives. We conduct empirical experiments on real data extracted from TripAdvisor, the TripR-2020Large dataset. \ph{The analysis of} results show a promising branch for developing quality decision making models using ChatGPT. Finally, we discuss \ph{the challenges of} consistency, sensitivity and explainability \ph{associated to the use of LLMs in CDM processes}, raising open \ph{questions} for future studies.
\end{abstract}

\section{Introduction}
Opinions are expressions of a person's beliefs, judgements, or attitudes about a particular topic or entity. They have a pivotal influence in our behaviors since our perceptions and choices change conditioned upon how others evaluate the world \cite{liu2012sentiment}. Consequently, it is common for us to actively seek the opinions of others when faced with a decision. \czm{Taking into account and analyzing the opinions of different people to make a decision can be exhausting and time consuming. It is convenient to have systems that contribute to an informed, efficient and objective decision process.} Decision \czm{Making (DM)} models are computational systems helping people \czm{to solve decision situations by analyzing the underlying opinions of experts who evaluate the potential alternative solutions.}

\rms{In recent years, there has been a trend towards the integration of opinions even from non-expert users in DM \cite{trillo2023group}. Furthermore, there has been a notable} increase in the number of users providing opinions, with Internet being a predominant source of knowledge \cite{viedma2021decision}. In this line, we can find many studies defining and expanding different methodologies, such as consensus \cite{9253574}, social network group \czm{DM} \cite{mengqi2023social} and large scale DM \cite{LIU2022220}, among others. \rms{Because opinions in \czm{social media platforms} are naturally expressed through \czm{plain texts},} crowd decision making (CDM)\cite{zuheros2022crowd} arises as a \czm{DM} model that leverages the wisdom of crowds \cite{surowiecki2005wisdom} \new{analyzing the opinion from free natural language texts, such as the reviews from social media platforms.} 

\new{Identifying the opinions from texts and inferring the evaluations of the experts to constitute the input of the DM model is challenging since they are integrated into the context and language nuances of the discourse. We define an evaluation as: \textit{the sentiment expression of an opinion that is linked to an alternative (or to a criterion of the alternative) from an expert review.} In order to distill the evaluations from the opinions expressed on the expert reviews, it} \czm{is convenient to incorporate Sentiment Analysis (SA) which is the branch of the Natural Language Processing (NLP) field concerned with the computational treatment of opinions and sentiments, and subjectivity in texts \cite{pang2008opinion}.} In this line, social media analytics leads to understand the pulse of the society through Internet forums, blogs, and posted reviews \cite{viedma2021decision}. \rms{Techniques} such as lexical-based approaches, machine learning, and deep learning models \cite{ren2018innovative, wen2019memristive, zuheros2022crowd} \rms{are used}.

Large Language Models (LLMs) have ushered a new \rms{route to \new{extract} opinions from} texts. They are statistical models with a substantial number of parameters that assign probabilities to sequences of words and are trained on a vast amount of data \cite{radford2019language, brants2007large}. They have a broad capacity to interpret and \rms{to} generate natural language \rms{texts}. Some prominent LLMs are LLaMa 2, Alpaca, LaMDA, Claude, \czm{PaLM 2}, Gopher, Falcon, StarCoder, Cohere, GPT-3.5, GPT-4 \czm{and Gemini} among others. \rms{The current most widely used --with extraordinary success in multiple \czm{NLP} tasks-- is ChatGPT~\cite{openai2021chatgpt}, a specific implementation for conversational purposes of GPT-3.5.}

\ph{We hypothesize on the incorporation of LLMs in CDM} \ph{based on prompt design strategies. We consider decision problems where experts offer opinions  though free texts in natural language. We integrate LLMs into CDM models since LLMs offer a deep semantic understanding of the language, seeking an \new{evaluation} of opinions by means of different types of scoring, based on different strategies of final representation of the score, numerical or linguistic, and taking into account different possibilities in the use or not of criteria and the individual or joint treatment of \new{reviews}. Using LLMs and prompt design strategies would be  possible to exploit the wisdom of the crowd from social opinions and make decisions.}

In this context, under these hypotheses, the aim of this paper is focused on the following three objectives and tasks:
\begin{itemize}
    \item {\it To integrate LLMs in \czm{\czm{DM} models}}. The goal is to incorporate  ChatGPT into CDM \czm{processes} that \rms{exploit} the wisdom of the crowd \ph{from social opinions by analyzing a massive number of free texts in natural language}. \czm{First,} we propose a workflow introducing the steps for CDM processes using ChatGPT. \czm{Secondly, we present multiple \ph{models} to infer and manage the \new{evaluations} from the \new{expert's reviews} by using ChatGPT. \ph{The first three models ask ChatGPT to extract an evaluation from each expert's opinions of an alternative}. The \ph{evaluations} are requested as sentiment polarity ratings, numerical ratings, or linguistic evaluations through the use of a linguistic terms set. The fourth approach conducts multi-criteria decision marking using a category ontology to consider criteria. The fifth scenario explores the use of ChatGPT as an end-to-end \czm{DM} model.} \rms{It is prompted to provide a ranking of the best alternatives based on the rates given by ChatGPT itself. \czm{We present prompt design strategies to adapt ChatGPT to the diverse previous approaches.}} 
    
    \item {\it \czm{To solve a case study with real data applying the CDM based on LLMs approaches}.} \czm{We conduct an experimental analysis to evaluate the feasibility of the previous theoretical proposals. In the five scenarios we} consider the TripR-2020Large dataset \cite{zuheros2022crowd}. It contains real \new{opinions} from \czm{the social media platform} TripAdvisor over four restaurants as plain natural language texts (132 users rate 4 restaurants located in London, \czm{providing} a total of 2586 \czm{opinions expressed in 474 \new{reviews}}). Results show a promising branch for developing quality \czm{DM} models using ChatGPT. \ph{This case of study will allow us to provide a deep analysis on our hypotheses.}
    
    \item {\it To open a discussion on three \ph{challenges:} consistency, sensitivity and explainability}. \rms{There is a need to design and guarantee computational models with these properties. This is why we analyze ChatGPT both as a complement to CDM and as a stand-alone solution.}
\end{itemize}
 
This paper is structured as follows. Section \ref{sec2} exposes an analysis about natural language in \czm{DM}, the CDM process, and a short analysis on the ChatGPT use and considerations on the prompts design. Section \ref{sec3} \czm{proposes} the ChatGPT based CDM models to \czm{solve decisions problems in five different scenarios}. Section \ref{sec_case_study} \czm{presents an experimental study conducted on the real data from the TripR-2020Large dataset}. Section \ref{sec4} offers the discussion on  consistency, sensitivity and explainability, raising some LLMs challenges based \czm{DM} models. Section \ref{sec5} provides some concluding remarks.

\section{Preliminars}\label{sec2}

\subsection{Natural language in multi-criteria decision making}\label{sec_21}
\new{Multi-criteria DM (McDM)} \rms{models are able to generate a ranking of best alternatives when they use the assessments of experts \cite{triantaphyllou2000multi}. Experts are requested to assess each alternative according to different attributes or criteria.} The most natural way for experts to evaluate entities, products, or any other type of alternatives is to use natural language since it aligns with the way people instinctively communicate and express their thoughts. Thus, the use of natural language assessments can benefit McDM models \cite{losee2001natural}.

One of the initial studies integrating free natural language \new{assessments} into McDM proposed the SA-MpMcDM methodology \cite{zuheros2021sentiment}. Its workflow handles the information in a comfortable way \new{through three phases.} The first phase captures or collects the expert \new{assessments}, which are natural language texts without restrictions. The second phase extracts the opinions from the \new{assessments} by means of sentiment analysis models. The third phase aggregates them and conducts the exploitation step getting the final ranking.


There is an emerging trend of studies that integrate natural language in McDM. 
Cao \textit{et al.} \cite{cao2022risky} present a risky large group emergency DM process based on topic sentiment analysis applied to public comments from network environments. Tayal \textit{et al.} \cite{tayal2023personalized} propose a McDM model that handles linguistic opinions and integrates customer preferences to get personalized ranking of products from e-commerce websites. Zhu \textit{et al.} \cite{zhu2023method} build an emergency DM method for the dynamic collaboration of the public and experts, based on social media data analyzed through a Bayesian estimation. \rms{In fields such as medicine, studies manifest} the importance of designing clinical natural language systems for healthcare DM \cite{hossain2023natural}. 

\subsection{Crowd decision making}\label{sec_22}
The wisdom of the crowd theory states that diverse, decentralized and independent groups of people are more intelligent than an elite few, no matter how expert they are \cite{surowiecki2005wisdom}. \czm{Social media platforms are current environments where people form this type of intelligent groups in a natural and disinterested way.} It is convenient to take advantage of the knowledge that these groups offer to design DM models that integrate different perspectives and a broader knowledge of the world.

Crowd Decision Making (CDM) is defined as a DM model that captures the wisdom of the crowd offered by large groups of people in natural environments \cite{zuheros2022crowd}. It has the advantage that there is no need to hunt for great experts to solve the decision task. The individuals from the large group constitute the traditional experts. Actually, it relies on the SA-MpMcDM methodology, since it process unrestricted texts, and extends it to large scenarios that profits the wisdom of crowds. 

CDM models are intended to help people solve real decision problems, so they should be evaluated on real data. A dataset, created from real \rms{reviews} published in TripAdvisor \cite{valdivia2017}, allows evaluating \czm{DM} models that integrate real natural language assessments. \rms{The TripR-2020Large dataset\footnote{\url{https://github.com/ari-dasci/OD-TripR-2020Large}} contains 474 written reviews in English of 4 London restaurants, provided by a total of 132 users. It is not complete as some users only evaluated 3 alternatives.} The unconstrained texts of the reviews are manually labelled by three annotators identifying a total of 2586 opinions. Each opinion has:
\begin{enumerate*}[label=(\arabic*)] 
\item a target \rms{alternative}, that identifies the object of opinion, 
\item a category, that identifies the criteria to which the target element belongs, and 
\item a polarity, that provides the negative, neutral or positive sentiment. 
\end{enumerate*} The set of available criteria is $C = \{$\textit{restaurant}, \textit{service}, \textit{food}, \textit{ambience}, \textit{drinks}, \textit{location}$\}$. 

\subsection{ChatGPT use and impact. Prompts design}\label{sec_23}
Created by OpenAI, ChatGPT is an advanced language model to generate human-like responses and engage in natural language conversations. It encompasses a wide range of topics and a vast amount of knowledge and information to understand and produce text in an conversational manner. ChatGPT is based on the Generative Pre-trained Transformer architecture GPT-3.5. \rms{One of the things that has made ChatGPT so successful with the general public is its user-friendly interface,} which allows seamless and accessible interactions for individuals without requiring technical expertise. \rms{The most common queries, according to ChatGPT itself, are} writing assistance, content generation, conversational agents, personalized recommendations, and problem-solving.  

The impact of ChatGPT within the scientific community is also highly remarkable. Zhao \textit{et al.} \cite{zhao2023chatagri} propose ChatAgri as a ChatGPT-based approach for agricultural text classification. Kocón \textit{et al.} \cite{kocon2023chatgpt} examine the capabilities of ChatGPT in multiple natural language tasks including \rms{SA}. Maddigan \textit{et al.} \cite{maddigan2023chat2vis} design the Chat2VIS system to convert natural language into code for appropriate visualisations using ChatGPT among other models. \ph{Sorin \textit{et al.} \cite{sorin2023large}  evaluate ChatGPT as a support tool for breast tumor board DM, asking the chatbot to recommend management.   Amin \textit{et al.} \cite{Ami2023} explore the fusion capabilities of ChatGPT with traditional NLP models in affective computing problems. Yang \textit{et al.} 
\cite{yang2023autogpt} present a benchmark study of Auto-GPT, an autonomous agent inspired by ChatGPT, in online DM tasks 
 introducing the ``Additional Opinions algorithm", as an easy and effective method that incorporates supervised/imitation-based learners into the Auto-GPT scheme. The aim of this approach is enabling lightweight supervised learning without requiring fine-tuning of the foundational LLMs.}
The use of ChatGPT \ph{is continuously growing,} \czm{applied} to many different areas such as computational social systems \cite{wang2023chatgpt}, intelligent vehicles \cite{gao2023chat}, art \cite{guo2023can}, and education \cite{nikolic2023chatgpt}, among others.

\begin{figure}[!th]
    \centering
    \includegraphics[width=\linewidth]{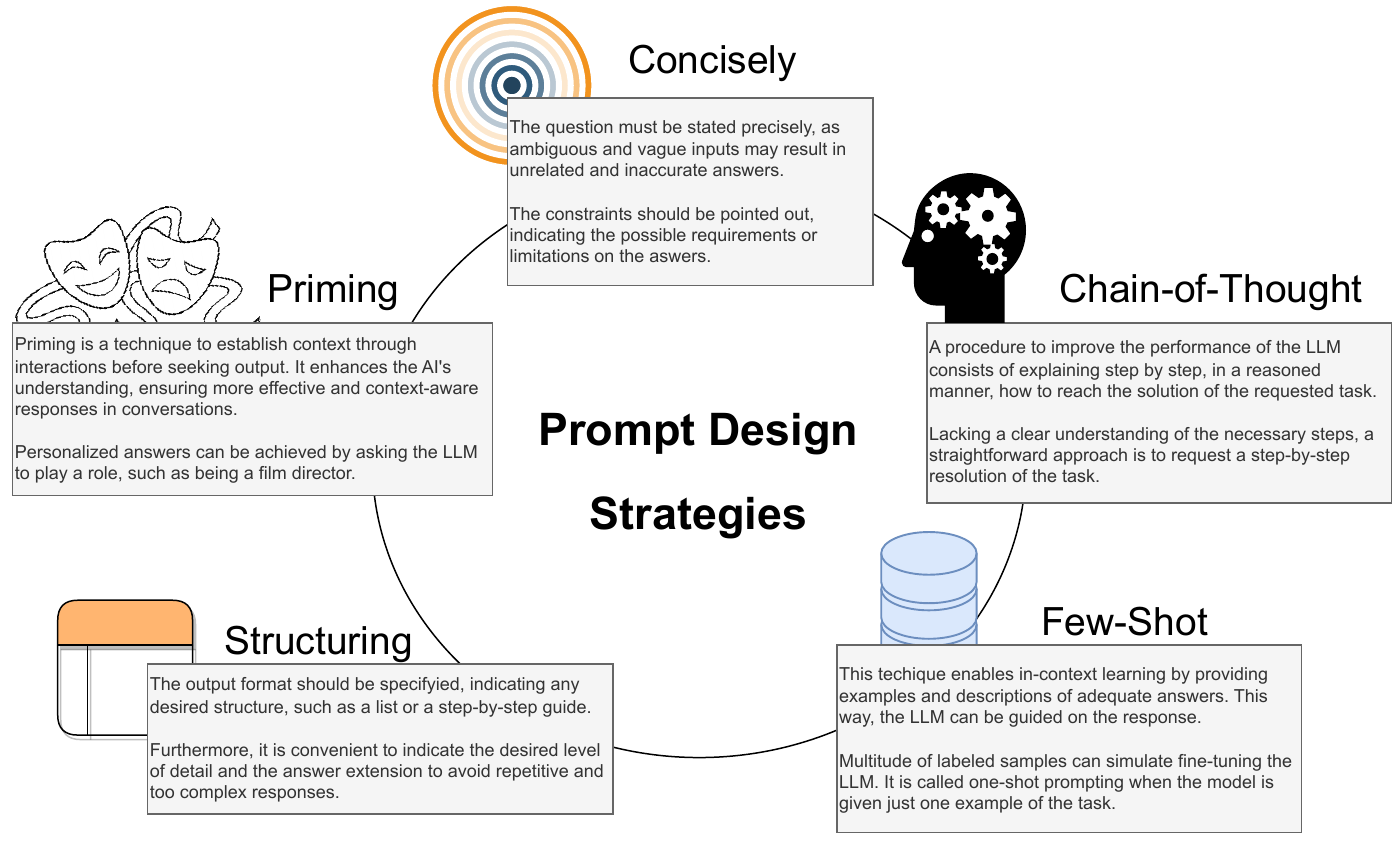}
    \caption{\czm{Prompt design strategies.}}
    \label{fig:promptdesign}
\end{figure}

Prompts are instructions given to LLMs to enforce rules, automate processes, and ensure specific qualities (and quantities) of generated output. They are also a form of programming to customize the outputs and interactions with LLMs \cite{white2023prompt}. Prompts are key elements to empower ChatGPT responses since the way a question is formulated influences the answer. \czm{Figure \ref{fig:promptdesign} gives} some tips to create effective prompts.

\section{ChatGPT based \czm{crowd} decision making models}\label{sec3}

\czm{This section presents how to integrate LLMs, such as ChatGPT, into \czm{DM} models that leverage the wisdom of crowds from social media platforms. We consider the experts are users who publish their reviews on a set of alternatives of interest \new{so, we particularize the reviews as expert's reviews.  We call evaluation to the sentiment expression of an opinion that is linked to an alternative, or criterion of the alternative, from an expert review. The expression of the sentiment can be extracted in terms of polarity, numerical values or by linguistic expressions. In the context of this work, the process of distilling the sentiment from an opinion provides an evaluation. Thus, we distill evaluations from the opinions of expert reviews. To do that}, the models require handling unrestricted natural language texts. We propose five scenarios to extract meaningful \ph{evaluations} \new{from the opinions expressed on the expert reviews} and to conduct the decision process by using ChatGPT:}
\begin{enumerate}
    \item Scenario with primary polarities.
    \item Scenario with numerical scores.
    \item Scenario with linguistic terms set.
    \item Scenario multi-criteria with \czm{a} category ontology.  
    \item \czm{Scenario end-to-end CDM system.}
\end{enumerate}

\subsection{Workflow of Crowd Decision Making based ChatGPT}
\begin{figure}[!ht]
    \centering
    \includegraphics[width=\linewidth]{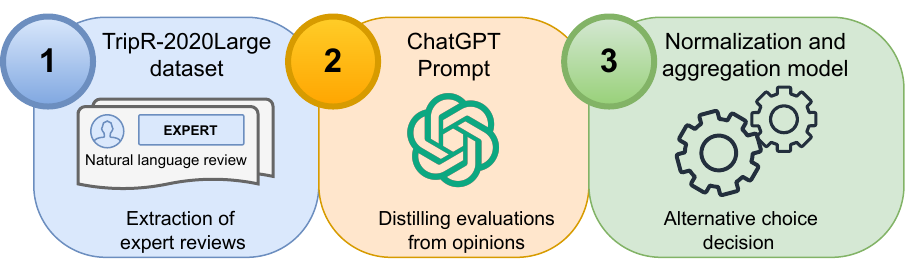}
    \caption{Workflow of CDM models based on ChatGPT.}
    \label{workflow_scenarios}
\end{figure}

\czm{The workflow followed in all the scenarios} is shown in Figure \ref{workflow_scenarios}\czm{, the phases of which are briefly presented below}. 
\begin{itemize}
    \item \new{Extraction of expert reviews. This phase} \czm{is identical across all scenarios. We define a CDM problem and a set of alternatives $X=\{x_1, \dots, x_n\}$ as potential solutions. The alternatives can encompass retail products, urban accommodations, or even political entities that promote their presence and offerings in social media. This phase downloads the actual reviews that users posts about the alternatives of interest. Users, who are the DM experts, are not forced to assess all the alternatives. The plain text from each published review is denoted by [DOCUMENT] and analyzed in the following step.} 

    \item \new{Distilling evaluations from opinions. This phase} consists of designing quality ChatGPT prompts in order to extract the \czm{sentiment or polarity of the opinions within the plain texts provided by the experts. The \ph{evaluation} \new{is the sentiment expression of an opinion linked to an alternative (or to a criterion of the alternative) from an expert review. It} can be represented by primary polarities (positive, negative or neutral), numerical scores (using the unit interval) or linguistic term sets (using natural language terms). Furthermore, the \new{evaluations} can be distilled at different levels of detail: 
    \begin{enumerate*}[label=(\arabic*)] 
    \item requesting the overall sentiment expressed on each alternative in each document,
    \item requesting the specific sentiment expressed on particular criteria of the alternatives in each document, or
    \item requesting the overall sentiment of each alternative in all associated documents.
    \end{enumerate*} We conduct the first level of detail using the three kind of representations, and the remaining levels of detail using just one kind of representation to avoid redundancy, giving rise to the five proposed scenarios.  }

    \item Alternative choice decision. It generates the ranking of the \czm{alternatives. In the initial four scenarios, it aggregates the \new{evaluations} identified by ChatGPT in each review. The \new{evaluations}} are offered in different formats and may require an answer normalization. \czm{In the last scenario, we order the overall \new{evaluations} indicated by ChatGPT about each alternative based on all \new{expert's} reviews so there is no need for normalization nor aggregation.} 
\end{itemize}

\czm{The following subsections present in detail the peculiarities of the second and third phases for each proposed scenario, paying special attention to the ChatGPT prompts and the normalization and aggregation of the models, respectively.}

\subsection{Decision making scenario with primary polarities}\label{sce_3labels}
This scenario distils the \new{evaluations} by asking ChatGPT the sentiment of the reviews as \textit{negative}, \textit{neutral}, or \textit{positive}, and computes the final ranking by the positive relative frequency.  

\paragraph{ChatGPT Prompt}
We request ChatGPT to indicate the predominant sentiment of each review at document level as a traditional negative, neutral, or positive polarity term. ChatGPT tries to offer a complex phrase as a response by default. Therefore, we force it to provide a single label in order to facilitate the output processing and \czm{to} avoid a complex answer normalization. The prompt is designed as follows:
\chatbox{Classify the sentiment of the following text as positive, neutral or negative, the answer must be a single label and one word: \newline [DOCUMENT]}

\paragraph{Normalization and aggregation model} The output obtained for each review corresponds to one single label: \textit{negative}, \textit{neutral}, or \textit{positive}. The final polarity rating associated to the \czm{alternative $x_i$, $i=1,\dots, n$} is computed as the positive relative frequency, \textit{i.e.}, $\czm{PolRating(x_i)}$ = 
$\frac{\#Positive(x_i)}{Total(x_i)}$ \czm{where $\#Positive(x_i)$ is the sum of reviews evaluating $x_i$ with positive polarity according to ChatGPT and $Total(x_i)$ is the number of reviews evaluating such alternative}.

\subsection{Decision making scenario with numerical scores}\label{sce_num}
This scenario \new{extracts meaningful information from the expert reviews} by asking ChatGPT \ph{to distil the evaluation} as a numeric rating in \czm{the interval} $[0,1]$. It computes the final ranking by averaging the \czm{previous numerical scores}.  

\paragraph{ChatGPT Prompt} We request ChatGPT to provide the polarity expressed within the interval $[0,1]$ for each document review. Values close to zero mean that a predominantly negative sentiment is expressed while values close to ones refer to a positive sentiment. The prompt is designed as follows:
\chatbox{Classify the sentiment of the following text using a score between 0 and 1, where 0 represents a completely negative sentiment and 1 represents a completely positive sentiment. The answer must be only a number: \newline [DOCUMENT]}

\paragraph{Normalization and aggregation model} The previous prompt forces ChatGPT to provide a single number, so no answer normalization is required. We collect the scores obtained for the reviews associated to the \czm{alternative} $x_i$ into the array $Scores(x_i)$. Then, we compute the average value of the score outputs associated with each \czm{alternative} getting the aggregated numerical rating\czm{, \textit{i.e.},} $\czm{NumRating(x_i)} = \overline{Scores(x_i)}$.

\subsection{Decision making scenario with linguistic terms set}\label{sce_lts}
This scenario asks ChatGPT to distil the \ph{evaluations from the reviews as a linguistic variable on} a linguistic term set $S=\{s_0, \dots, s_g\}$ with $g+1$ odd cardinality. The final ranking is computed by the 2-tuple weighted average \cite{martinez20152}. 

\paragraph{ChatGPT Prompt} We force ChatGPT to provide the polarity as a five-cardinality linguistic scale since usually some texts manifest strongly positive (or negative) sentiment while others express only predominantly positive (or negative) sentiment. Thus, we establish the linguistic term set $S=\{$Very Negative, Negative, Neutral, Positive, Very Positive$\}$. The prompt is designed as follows:
\chatbox{Classify the sentiment of the following text as very positive, positive, neutral, negative or very negative, your answer must only be the label: \newline [DOCUMENT]}

\paragraph{Normalization and aggregation model} ChatGPT offers a unique word for each review specifying its polarity in a five linguistic scale. We design the aggregation model by means of the 2-tuple linguistic model in order to avoid the loss of information \cite{martinez20152}. Let $\overline{S} = S \times [-0.5, 0.5)$ be the 2-tuple set associated with $S$. The function $\Delta_s:[0,g] \rightarrow \overline{S}$ is given by 
\begin{equation}
    \Delta_{S}(\beta) = (s_i, \alpha) \; with \left\{
			\begin{array}{ll}
				i = round(\beta) \\
				\alpha = \beta - i \\
			\end{array}
			\right.
\end{equation}
Note there is a function $\Delta^{-1}_{S}(s_i, \alpha) = i + \alpha$.

We compute the aggregated 2-tuple rating $\czm{2TRating(x_i)}$ associated to the \czm{alternative $x_{i}, i=1, \dots, n$} by a simplification of the 2-tuple weighed average as shown in Equation \ref{eq:agg2tuple}.
\begin{equation}\label{eq:agg2tuple}
     \czm{2TRating(x_i)} =  \Delta_{S}\left(\dfrac{\sum_{i=0}^{g}{i}\times\omega_{i}}{\sum_{i=0}^{g}\omega_{i}}\right)
\end{equation}
where \czm{$\omega_{i} = \{\#$\textit{V.Neg.}$(x_i)$, $\#$\textit{Neg.}$(x_i)$, $\#$\textit{Neutral}$(x_i)$, $\#$\textit{Pos.}$(x_i)$, $\#$\textit{V.Pos}$(x_i)$\} is} the weight associated to $x_i$. 

The final rating is computed by normalizing the 2-tuple ratings to the unit interval, \textit{i.e.}, $\czm{Normalized2TRating(x_i)} = \frac{\Delta^{-1}_{S}(\czm{2TRating(x_i)})}{g}$.

\subsection{Multi-criteria decision making scenario with a category ontology}\label{sce_mc}
This scenario requests ChatGPT the \ph{evaluation} for criteria using a category ontology associated with the criteria aspects. We \czm{represent the distilled \new{evaluations} by the primary polarities}  \textit{negative}, \textit{neutral}, and \textit{positive}. Then, we compute the positive relative frequency for each category and get the mean value to set the final ranking.

\paragraph{ChatGPT Prompt} We request ChatGPT for a sentiment analysis of the reviews where it must identify the \textit{negative}, \textit{neutral}, or \textit{positive} polarity expressed about multiple categories. \czm{A category represents a set of similar but not necessarily equivalent opinion targets of an alternative or entity. For example, supposing the alternatives are restaurants, the category \textit{food} can be designated to group targets such as salad or lemon cake. Moreover, the categories can be evaluated regarding different attributes to retain more insights regarding the opinions. For example, the category \textit{food} may have attributes such as price and quality. We call \textit{category ontology} to the decomposition of alternatives into category-attribute pairs and represent it by \{$Category_1$\#$Attribute_1$, $\dots$, $Category_m$\#$Attribute_{m'}$\}. It is possible to dispense with attributes when a less concise level of detail is deemed sufficient. In any case, this set can be considered as detailed criteria of the alternatives so this scenario actually solves a McDM problem.} Some reviews may not offer opinions on some \czm{category pairs}, so we request ChatGPT to set sentiment \textit{None} in such cases. The prompt is designed as follows: 

\chatbox{Classify the sentiment of the following text, evaluating the following categories: $Category_1$\#$Attribute_1$, $\dots$, $Category_{m-1}$\#$Attribute_{m'-1}$ and $Category_{m}$\#$Attribute_{m'}$. If there is no specific mention or evaluation of a category in the text the label should be None, while the categories that are mentioned should be labeled with {positive, neutral or negative}. The format of the answer should be \small{`CATEGORY: LABEL'} where \small{CATEGORY} should be replaced with one of the categories to evaluate and \small{LABEL} with the evaluation's result:\newline [DOCUMENT]}

\paragraph{Normalization and aggregation model} It is necessary a normalization of the output provided by ChatGPT since is not limited to a single word or number. 
We must extract the \czm{evaluated categories} and their associated sentiments. Then, for each alternative, we count the number of \textit{Positive}, \textit{Negative}, \textit{Neutral}, and \textit{None} sentiment associated with each category. Afterwards, we compute the positive relative frequency ignoring the \textit{None} values for all the categories. \czm{Finally, we get the multi-criteria polarity rating $MCRating(x_i)$ value as the positive relative frequency for each alternative (considering all criteria are equally important).}

\subsection{\czm{ChatGPT as an end-to-end decision making system}}\label{sce_endtoend}
\czm{This scenario explores} the use of ChatGPT as a global CDM system, an end-to-end \czm{DM} system \ph{that considers natural language plain} \new{expert reviews}. We request ChatGPT to play the role of a decision maker \czm{who must analyze simultaneously all the reviews associated to each alternative and provide the overall \new{evaluation} as a numerical score. Then, the final ranking is obtained by ordering such scores.}

\paragraph{ChatGPT Prompt} \czm{In this scenario, the plain text [DOCUMENT] compiles all the expert's reviews that evaluate an alternative. We feed ChatGPT with them and make two requests: a general opinion and a general score. The first one attempts to provide a general summary of all the reviews while the second one leads to the alternatives ranking. Actually, the first question forces ChatGPT to completely revise the input texts in order to provide a more confident general score in the second question. The prompt is designed as follows:}

\chatbox{[DOCUMENT]\footnote{In this scenario, DOCUMENT is a set containing all the expert's reviews that evaluate an alternative.}\\
What's the general opinion about the alternative? \\
Then, what’s the general sentiment score about the alternative? The score must be a number between 0 and 1, your answer must only be a number.}

\paragraph{Normalization and aggregation model} \czm{The numerical \new{evaluation} that ChatGPT provides at the end of its response, \textit{i.e.} the $ChatGPTRating$ value, is compiled for each alternative. Then, we order them to obtain the final ranking. }

\section{\czm{Case study\ph{: Restaurant ranking from TripAdvisor \new{reviews}} }}\label{sec_case_study}
\czm{In this section, we apply the proposed ChatGPT based \czm{CDM} models to a practical DM problem for choosing restaurants. We consider the TripR-2020Large dataset (see Section \ref{sec_22}) since it collects real data to evaluate DM models with unrestricted natural language inputs. It contains the actual reviews from 132 TripAdvisor users. They express their \czm{experiences} on four restaurants, which constitute the set of alternatives:} $X = \{x_{1}, x_{2}, x_{3}, x_{4}\}$ = $\{${\textit{The Oxo Tower}}, {\textit{The Wolseley}}, {\textit{The Ivy}}, {\textit{J. Sheekey}}$\}$. Not all experts \new{review} all alternatives, so there are slightly less than $132\times4$ documents. \czm{Hereunder, we present how to solve the CDM problem for selecting the best restaurant in our five proposed scenarios by incorporating ChatGPT.} \ph{Finally, a deep comparative analysis is presented, which serves to introduce some challenges. }

\subsection{Scenario with primary polarities}\label{cs_3labels}
\begin{figure}[!hbt]
    \centering
    \includegraphics[width=\linewidth]{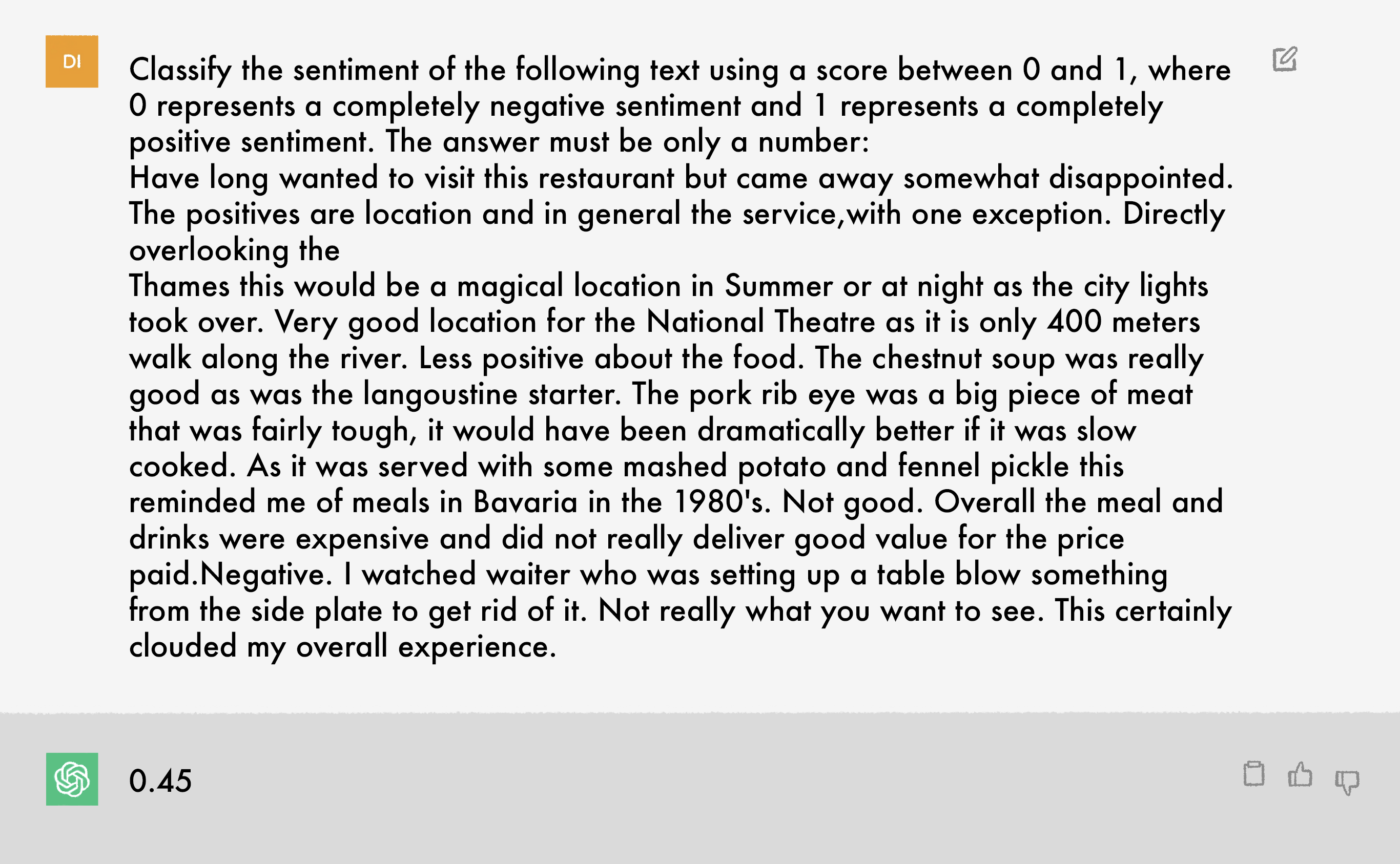}
    \caption{ChatGPT result for classifying \new{evaluation} as a positive, neutral, or negative polarity.}
    \label{fig:gpt-basic}
\end{figure}

\czm{We conduct the model exposed in Section \ref{sce_3labels}. First, we request ChatGPT to distill the \new{evaluations} for each review collected in the TripR-2020Large dataset as a negative, neutral, or positive polarity. For example, Figure \ref{fig:gpt-basic} shows the text offered by a user \new{reviewing} a restaurant and the distilled \new{evaluation} by ChatGPT, which is negative. This identified \new{evaluation} is quite suitable as the user begins by highlighting positive aspects of the restaurant but they are outweighed by the negative points that \rms{they detail later}. This scenario loses the sentiment nuances when assigning the predominant polarity.}

\czm{Once ChatGPT identifies the overall \new{evaluation} of each review, we conduct the normalization and aggregation process. }Table \ref{tab:freq-ext-1} shows the absolute frequency of each sentiment and the aggregated polarity rating for each restaurant.  Attending to such ratings, the final ranking is $x_{4} > x_{3} > x_{2} > x_{1}$, \textit{i.e.}, \textit{J. Sheekey} $>$ \textit{The Ivy} $>$ \textit{The Wolseley} $>$ \textit{The Oxo Tower}.
\begin{table}[t]
    \centering
    \begin{tabular}{ccccc}
         Restaurant & \#Negative & \#Neutral & \#Positive & \czm{PolRating} \\
         \hline
         The Oxo Tower & 3  & 18 & 74  & 0.7789 \\
         The Wolseley  & 26 & 2  & 106 & 0.791\\
         The Ivy       & 22 & 8  & 118 & 0.7973\\
         J. Sheekey    & 16  & 1  & 80 & 0.8247 \\
    \end{tabular}
    \caption{Final restaurant rating for the scenario \ref{cs_3labels}. \czm{PolRating} is the frequency of positives.}
    \label{tab:freq-ext-1}
\end{table}

\subsection{Scenario with numerical scores}\label{cs_num}
\begin{figure}[!hbt]
    \centering
    \includegraphics[width=\linewidth]{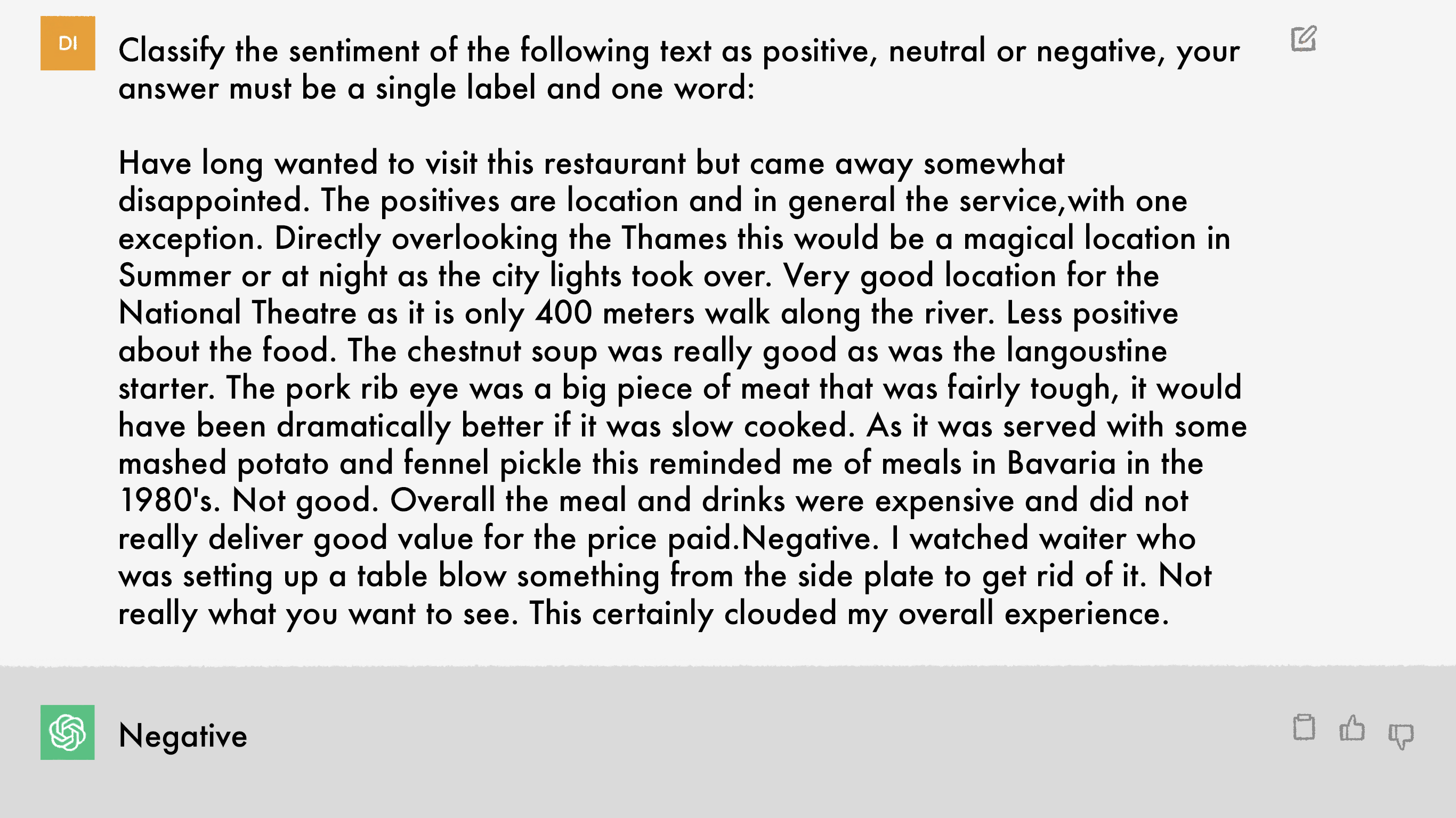}
    \caption{ChatGPT result for classifying \new{evaluation} as a numerical score.}
    \label{fig:gpt-score}
\end{figure}

\czm{We conduct the model exposed in Section \ref{sce_num}. Firstly, we request ChatGPT to distill the \new{evaluation} for each review of TripR-2020Large as a numerical rating in the interval $[0,1]$. For example, Figure \ref{fig:gpt-score} shows the review analyzed in the previous scenario (see Figure \ref{fig:gpt-basic}) and the \new{evaluation} distilled by ChatGPT, which is 0,45. This scenario better capture the overall expressed sentiment by the user since it offers a predominantly negative value but hints at positive points.}

\czm{Then, we aggregate the distilled \new{evaluations} as numerical scores.} Table \ref{tab:res_sce_num} shows the numerical score rating obtained for each restaurant, getting the final ranking $x_{4} > x_{3} > x_{2} > x_{1}$, \textit{i.e.}, \textit{J. Sheekey} $>$ \textit{The Ivy} $>$ \textit{The Wolseley} $>$ \textit{The Oxo Tower}.
\begin{table}[!hbt]
    \centering
    \begin{tabular}{cc}
         Restaurant & \czm{NumRating} \\
         \hline
         The Oxo Tower & 0.7632 \\
         The Wolseley         & 0.7652 \\
         The Ivy              & 0.7957 \\
         J. Sheekey           & 0.7988 \\
    \end{tabular}
    \caption{Final restaurant rating for the scenario \ref{cs_num}. \czm{NumRating} is the mean value of the numerical scores.}
    \label{tab:res_sce_num}
\end{table}

\subsection{Scenario with linguistic terms set}\label{cs_lts}
\begin{figure}[!hbt]
    \centering
    \includegraphics[width=\linewidth]{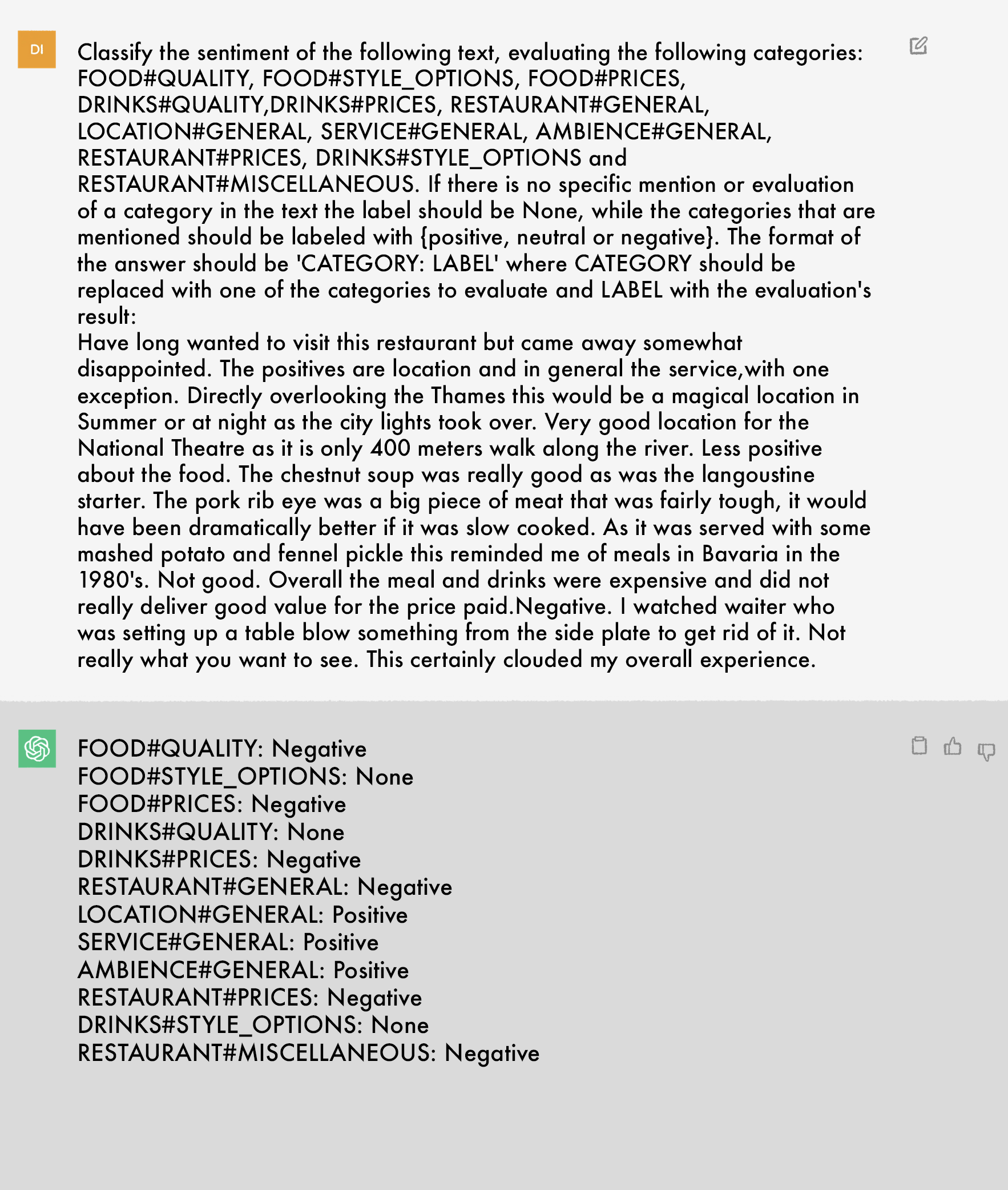}
    \caption{ChatGPT result for classifying \new{evaluation} as a very negative, negative, neutral, positive, or very positive polarity.}
    \label{fig:gpt-more-2}
\end{figure}

\czm{We conduct the model exposed in Section \ref{sce_lts}. We request ChatGPT to distill the \new{evaluations} from each review as very negative, negative, neutral, positive, or very positive. As an example, Figure \ref{fig:gpt-more-2} shows two reviews from TripR-2020Large. The first review is analyzed in the previous scenarios (see Figure \ref{fig:gpt-basic} and Figure \ref{fig:gpt-score}). In this scenario, ChatGPT manifests the \new{evaluation} is \textit{negative} which is: 
\begin{enumerate*}[label=(\arabic*)] 
\item a more fair relaxed negative sentiment than the one identified in the first scenario, and
\item more natural than providing the numerical rating $0,45\in[0,1]$ as in the second scenario. 
\end{enumerate*} The second review manifests a very positive opinion, as ChatGPT identifies.}

\czm{We conduct the normalization and aggregation process.} Table \ref{tab:freq-ext-3} indicates the number of reviews associated with each possible polarity for each restaurant. Positive and very positive \new{evaluations} predominate, leaving all the alternatives in an remarkably favorable position. \czm{Finally,} Table \ref{tab:res_sce_lts} shows the 2-tuple rating and the normalized 2-tuple rating obtained, which provides the final ranking $x_{4} > x_{3} > x_{2} > x_{1}$, \textit{i.e.}, \textit{J. Sheekey}$>$\textit{The Ivy}$>$\textit{The Wolseley}$>$\textit{The Oxo Tower}.
\begin{table}[t]
    \centering
    \begin{tabular}{ccccccc}
         Restaurant & \#V.Neg. & \#Neg. & \#Neutral & \#Pos. & \#V.Pos. \\
         \hline
         The Oxo Tower & 3  & 13 & 6  & 37 & 36  \\
         The Wolseley  & 2  & 19 & 7  & 61 & 45 \\
         The Ivy       & 1  & 20 & 12  & 62 & 53  \\
         J. Sheekey    & 1  & 11 & 4  & 43 & 38 \\
    \end{tabular}
    \caption{Sentiment frequency using the term set $S=\{$Very Negative, Negative, Neutral, Positive, Very Positive$\}$.}
    \label{tab:freq-ext-3}
\end{table}

\begin{table}[t]
    \centering
    \begin{tabular}{ccc}
         Restaurant & \czm{2TRating} & \czm{Normalized2TRating}\\
         \hline
         The Oxo Tower & ($s_3$, -0.0526) & 0.7369\\
         The Wolseley         & ($s_3$, -0.0448) & 0.7388\\
         The Ivy              & ($s_3$, -0.0135) & 0.7466\\
         J. Sheekey           & ($s_3$, 0.0928) & 0.7732\\
    \end{tabular}
    \caption{Final restaurant rating for the scenario \ref{cs_lts}. \czm{Normalized2TRating} is the normalized 2-tuple weighted average associated to $S$ \czm{and} weights are the sentiments frequency.}
    \label{tab:res_sce_lts}
\end{table}

\subsection{Scenario with a category ontology}\label{cs_mc}
\czm{We conduct the model exposed in Section \ref{sce_mc}. Thus, we solve a McDM problem where we ask ChatGPT to identify the expressed sentiments on reviews for particular criteria associated to the restaurants.} Figure \ref{fig_ontology} presents the \czm{proposed restaurant} category ontology that is made up of 12 \czm{category-attribute pairs}. The root is the evaluated restaurant, the \czm{second} level consists of the criteria from the TripR-2020Large dataset, and the \czm{third} level correspond to the evaluable attributes of the criteria which are also indicated in the dataset.
\begin{centering}
\input{ontology}
\end{centering}

\begin{figure}[t]
    \centering
    \includegraphics[width=0.9\linewidth]{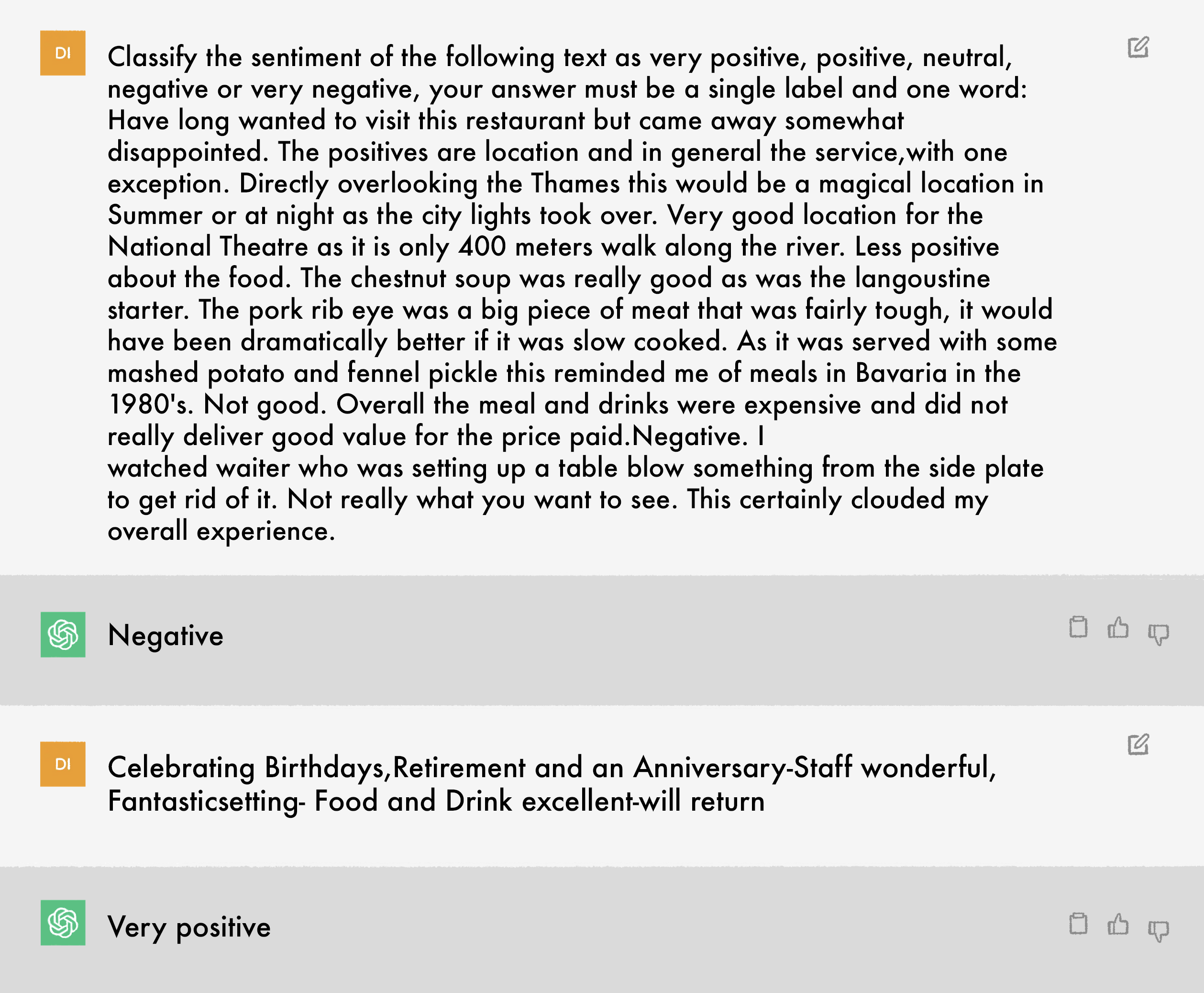}
    \caption{ChatGPT result for classifying \new{evaluation} as negative, neutral, or positive polarity for the category ontology elements.}
    \label{fig:gpt-cat}
\end{figure}

\czm{As an example,} Figure \ref{fig:gpt-cat} shows the \new{evaluation} \czm{identified} by ChatGPT \czm{for each} element of the ontology for the \czm{\new{expert review} analyzed in the previous scenarios (see Figure \ref{fig:gpt-basic}, Figure \ref{fig:gpt-score}, and Figure \ref{fig:gpt-more-2}). ChatGPT manifests that the \textit{negative} sentiment predominates, as the previous scenarios also pointed out. Successfully, this scenario allows to capture more expressed sentiments. It identifies positive \new{evaluations} for certain categories such as \textit{location} and \textit{service} (note the user specifies \textit{``The positives are location and in general the service"}).} 

\newcommand{\refappendix}[1]{\hyperref[#1]{Appendix~\ref*{#1}}}

\czm{The normalization and aggregation model counts the number of \textit{Positive}, \textit{Negative}, \textit{Neutral}, and \textit{None} \new{evaluations} for each category and computes its positive relative frequency.} This information is shown in the \refappendix{appendix:tables_MC} since is too extensive. The most relevant statistic, \textit{i.e.} the final \czm{positive} percentage value, is shown in Table \ref{tab:freq-multicriteria} as the multi-criteria polarity rating $\czm{MCRating}$ in the unit interval. Ordering such ratings, we get the final ranking $x_{4} > x_{1} > x_{2} > x_{3}$, \textit{i.e.}, \textit{J. Sheekey}$>$\textit{The Oxo Tower}$>$\textit{The Wolseley}$>$\textit{The Ivy}.

\begin{table}[t]
    \centering
    \begin{tabular}{ccccccc}
         Restaurant & \czm{MCRating}\\
         \hline
         The Oxo Tower &  0.681  \\
         The Wolseley  & 0.6526  \\
         The Ivy       & 0.622   \\
         J. Sheekey    & 0.7062  \\
    \end{tabular}
    \caption{Final restaurant rating for the scenario \ref{cs_mc}. \czm{MCRating} is the mean score of the positive percentage values associated to the categories (see \refappendix{appendix:tables_MC}).}
    \label{tab:freq-multicriteria}
\end{table}

\subsection{Scenario end-to-end decision making system}\label{cs_endtoend}
\czm{We use the model exposed in Section \ref{sce_endtoend} so we consider ChatGPT as an entire CDM system. It must analyze, as a decision maker,} the restaurants collected in the TripR-2020Large dataset. \czm{Therefore, ChatGPT receives as input a plain text [DOCUMENT] for each restaurant gathering all their \new{expert's reviews}. We request ChatGPT a general \new{evaluation} and a general sentiment score about each restaurant. Below we present the answer to first part of the prompt, which is the question \textit{``What's the general opinion about the restaurant?''}}
\begin{enumerate}
    \item \textit{\textbf{Oxo Tower Restaurant, Bar and Brasserie:} The general opinion about Oxo Tower Restaurant, Bar, and Brasserie is mixed. Some people rave about the stunning views, excellent service, and delicious food, while others have found the food overpriced and mediocre. The atmosphere, especially in the bar and brasserie areas, is often described as vibrant and lively. Overall, it seems to be a popular destination for special occasions and enjoying the iconic London skyline.}
    \item \textit{\textbf{The Wolseley:} The general opinion about The Wolseley is quite positive. Many people appreciate the elegant and grand atmosphere of the restaurant, the attentive service, and the range of food options available. Some reviewers mention the impressive interior and historic significance of the building, while others enjoy the buzzing atmosphere and the possibility of celebrity sightings. However, there are also some mixed opinions regarding the food quality, pricing, and occasional disappointing experiences with service. Overall, The Wolseley is often recommended as a popular option for a stylish and enjoyable dining experience in London's West End.}
    \item \textit{\textbf{The Ivy:} The general opinion about The Ivy is quite positive. Many people praise its classic and elegant atmosphere, attentive service, and good quality food. The restaurant is often recognized as an iconic and popular dining destination in London. However, there are also some mixed reviews regarding aspects such as the price point, overcrowding, or occasional inconsistency in the food. Overall, it seems that The Ivy still maintains a strong reputation and continues to attract a wide range of diners.}
    \item \textit{\textbf{J.Sheekey:} The general opinion about J. Sheekey is positive. Many people praise the restaurant for its fresh seafood, excellent service, and pleasant atmosphere. It is often recommended for pre-theatre meals, special occasions, and seafood lovers. However, there are also some mixed reviews, with a few customers expressing disappointment in certain aspects such as high prices, crowded seating, or inconsistent quality. Overall, it is considered a reputable seafood restaurant in London.}
\end{enumerate}

Finally, Table \ref{tab:freq-chatgptscores} shows the ratings provided by ChatGPT for all the restaurants. Ordering the \czm{ChatGPTRating} values, we get the final ranking $x_{3} > x_{4} = x_{2} > x_{1}$, \textit{i.e.}, \textit{The Ivy}$>$\textit{J. Sheekey}=\textit{The Wolseley}$>$\textit{The Oxo Tower}.
\begin{table}[!hbt]
    \centering
    \begin{tabular}{cc}
         Restaurant & \czm{ChatGPTRating}\\
         \hline
         The Oxo Tower &  0.59  \\
         The Wolseley  & 0.75  \\
         The Ivy       & 0.8   \\
         J. Sheekey    & 0.75  \\
    \end{tabular}
    \caption{\czm{Final restaurant rating for the scenario \ref{cs_endtoend}. ChatGPTRating is the numerical score provided by ChatGPT. }}
    \label{tab:freq-chatgptscores}
\end{table}

\ph{This scenario achieves a different ranking from the previous ones. Throughout the following subsection we analyze in depth the results obtained. In the next section, we pay special attention to the sensitivity consistency and explainability challenges of ChatGPT for CDM that appear in this analysis.}

\subsection{Global results analysis}\label{result_analysis}
Table \ref{tab:rankings} compiles the rankings of the restaurants from the TripR-2020Large dataset reached in the scenarios of Section \ref{sec_case_study}. Overall, the results are similar which manifests the proposed models are quite consistent. Furthermore, we provide the rankings achieved by the CDM guided by the SA (CDM-SA) model \cite{zuheros2022crowd} and the Explainable CDM based on Subgroup Discovery and Attention Mechanisms (ECDM-SDAM) methodology \cite{zuheros2023decision} since both tackle the same case of study and serve as baselines. We point out three main findings. 

\begin{table}[!hbt]
    \centering
    \resizebox{0.9\textwidth}{!}{  
    \begin{tabular}{cccccc}
         \multirow{2}{*}{Scenario}& \multicolumn{4}{c}{Score in the unit interval} & \multirow{2}{*}{Ranking}  \\
          & $x_1$ & $x_2$ & $x_3$ & $x_4$ &  \\
         \hline
         Primary polarities (\ref{cs_3labels}) & 0.7789 & 0.791 & 0.7973 & 0.8247 &  $x_{4} > x_{3} > x_{2} > x_{1}$  \\
         Numerical scores  (\ref{cs_num})& 0.7632 & 0.7652 & 0.7957 & 0.7988 &  $x_{4} > x_{3} > x_{2} > x_{1}$\\
         Linguistic terms set  (\ref{cs_lts}) & 0.7369 & 0.7388 & 0.7466 & 0.7732 & $x_{4} > x_{3} > x_{2} > x_{1}$   \\
         Multi-Criteria with ontology   (\ref{cs_mc}) & 0.681 & 0.6526 & 0.622 & 0.7062 & $x_{4} > x_{1} > x_{2} > x_{3}$   \\
         ChatGPT as a CDM system   (\ref{cs_endtoend}) & 0.59 & 0.75 & 0.8 & 0.75 & $x_{3} > x_{4} = x_{2} > x_{1}$  \\
         \hline
         CDM-SA model \cite{zuheros2022crowd} & 0.59 & 0.61 & 0.71 & 0.69 & $x_{3} > x_{4} > x_{2} > x_{1}$  \\
         ECDM-SDAM methodology \cite{zuheros2023decision} & 0.868 & 0.881 & 0.912 & 0.91 &  $x_{3} > x_{4} > x_{2} > x_{1}$  \\
         \hline
    \end{tabular}
    }
    \caption{Final restaurant rankings from TripR-2020Large.}
    \label{tab:rankings}
\end{table}


\begin{enumerate}
\item Most of achieved rankings match with the baseline ranking obtained by the CDM-SA model and the ECDM-SDAM methodology. The main difference is the order between restaurants $x_3$ and $x_4$ in the top of rankings. This is because both have similar high quality scores and ChatGPT evaluations have the inherent uncertainty of converting a text into a single assessment. This fact can alter the evaluations order as we will see in the sensitivity analysis (next section). In \cite{valdivia2018} is analyzed the mentioned inherent uncertainty among differences  SA machines, \ph{ they present differences for getting polarity, proposing the cited paper} to use a consensus model for filtering neutrality. Therefore, a potential extension could be to perform multiple runs and compute the average in each scenario in order to achieve most robust rankings. 

\item  The ranking obtained by the multi-criteria with category ontology scenario is the most discrepant since the restaurants $x_1$ and $x_3$ change their position quite drastically with respect to the other rankings. This may be because this scenario considers all categories to be equally important, whereas in a real environment it is common that, for example, \textit{food} may be more relevant than other. It might be interesting to set categories weights through the attention of the experts \cite{zuheros2021sentiment}, paying attention to the most relevant aspects and their associated criteria. \ph{On the other hand, many \new{expert's reviews} do not provide evaluations for different criteria, and this lack of information reduces the opinions of the crowd to a small set of \new{evaluations} per criterion and, therefore, the use of many criteria loses overall robustness. The analysis based on criteria and ontologies therefore requires an in-depth analysis, most likely the need to learn weights for the criteria, and to set a more robust model in line with an end-to-end evaluation. On the other hand, it is obvious to mention that the use of criteria can provide explainability over the decision model, an important challenge to be discussed in the next section. }

\item The ranking obtained by ChatGPT as a CDM system situates the restaurant $x_3$ in a better position than $x_4$, which \czm{happens only} in this scenario. This may be because the reviews of $x_3$ contain slightly more positively nuanced opinions or terms with more positive connotations than $x_4$. The reviews present several individual opinions as individual sentences, with different polarities among them, and requires an internal global evaluation as a summary among them. We analyze the distilled evaluations in the next section via consistency versus sensitivity. 
\end{enumerate}

\section{Discussion on challenges: \new{c}onsistency, sensitivity and explainability}\label{sec4}
This section, based on the results shown in \czm{Section \ref{sec_case_study}}, focuses the attention on three questions: the consistency of the \new{evaluations} for different representations, the sensitivity of ChatGPT, and the explainability of the \czm{achieved} decisions. They outline some challenges for future studies, for improving the ChatGPT based \czm{DM} models, \czm{allowing more robust models.} 

\subsection{Consistency}
The results provided by ChatGPT present some inconsistencies that should be highlighted. In some reviews, it gets quite different \new{evaluations} depending on the requested output format. Mostly, it happens in the first two scenarios, where we ask ChatGPT to distil evaluations by means of the polarity \textit{positive}, \textit{negative}, or \textit{neutral}, or trough numerical evaluations in the unit interval. As we have mentioned, this may be to the inherent uncertainty of converting a text into a single assessment. \ph{In \cite{huang2023look} is presented an interesting analysis via
an exploratory study of uncertainty
measurement for LLMs. }

\ph{On the one hand, much has been written about ChatGPT's hallucinations, in some ways these inconsistencies could be associated with an inaccurate inference and could be considered as particular kind of hallucination if they present high differences when evaluate criteria. We define an hallucination versus uncertainty measurement in this context as a very discrepant sentiment classification when distilling the evaluations in different formats, presenting quite different opinions between positive and negative positions. On the other hand, a review including this characteristic can also be considered an inconsistent \new{review} which may need special consideration. }

\begin{figure}[!ht]
    \centering 
    \subfloat{
    \includegraphics[width=0.45\textwidth]{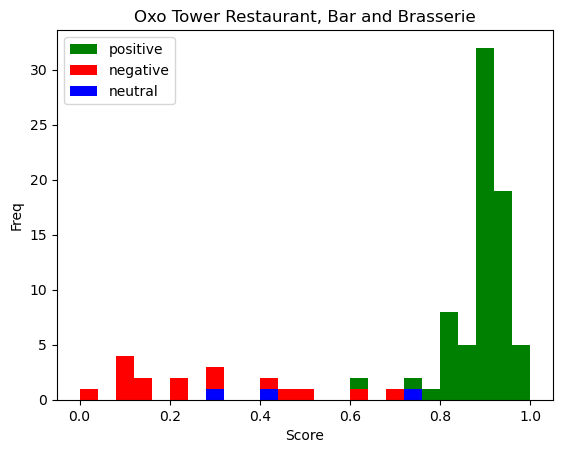}
    }
    \subfloat{
    \includegraphics[width=0.45\textwidth]{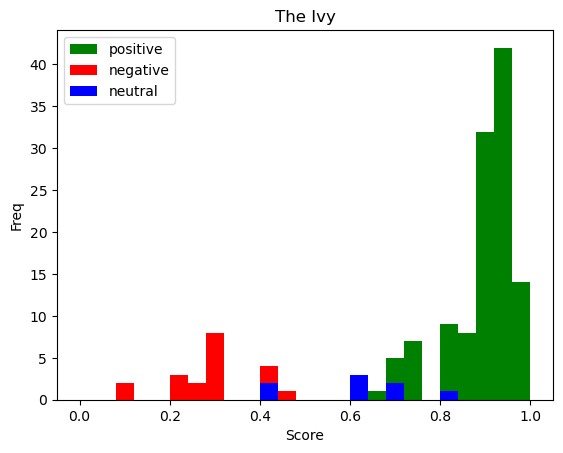}
    }
    \hfill
    \subfloat{
    \includegraphics[width=0.45\textwidth]{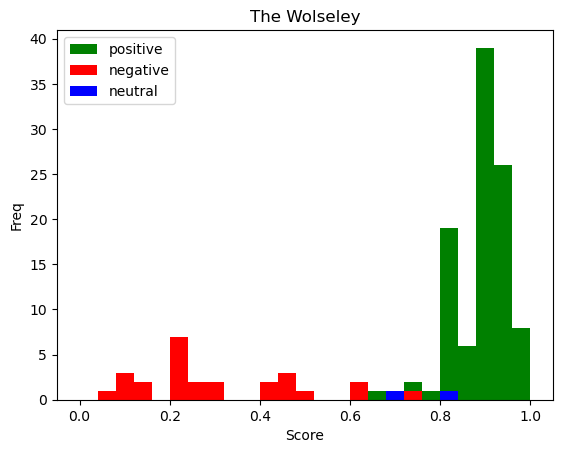}
    }
    \subfloat{
    \includegraphics[width=0.45\textwidth]{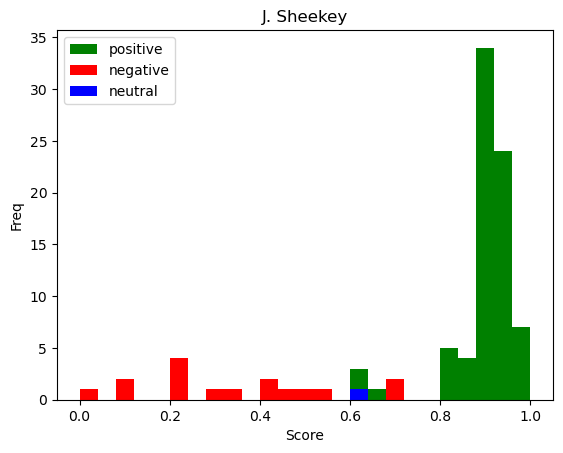}
    }
    \caption{ChatGPT inconsistencies for the four restaurants.}
    \label{fig:enter-label}
\end{figure}

Figure \ref{fig:enter-label} presents the inconsistencies generated for each evaluated restaurant. Visually, the inconsistencies correspond to the colors red and blue located far to the right, and to the colors green and blue located far to the left on the histograms. 

\begin{example}
The review shown in \czm{Figure \ref{review_hallucination}}, assessing The Oxo Tower restaurant, is analyzed by ChatGPT. There is an inconsistency since in the first scenario with primary polarities it outputs a \textit{negative} sentiment while in the second scenario with numerical scores it outputs a $0.7$ value. 

\begin{figure}
\examplebox{}
{We ate in the Restaurant rather than the Brasserie for a change and weren’t disappointed however, this is, as you would expect, one of those restaurant where less is more (apart from the constant supply of bread to fill you up!) The wine list is good, but not the best value but you are also paying for the view of the London skyline (although it was terrible weather therefore the view wasn’t that great unfortunately). It’s a bit too canteeny for our liking but we were glad we visited. This may not be a ringing endorsement but it’s not far from. As is often the case, trying to catch the waiter’s eye in order to get the bill want an easy task, it took ages...}
\caption{\czm{Review from TripR-2020Large in which ChatGPT produces an hallucination when inferring the main \new{evaluation}.}}
\label{review_hallucination}
\end{figure}
\end{example}

 It is an open issue to design ChatGPT based \czm{DM} models to manage inconsistencies or hallucinations dealing with the discussed inherent uncertainty in complex reviews. One possibility may be to consider the different single opinions from a review to reduce the uncertainty associated with the text in the context of a multi-criteria \czm{DM} model, \ph{with each single \czm{evaluation} associated to an aspect or criterion but paying attention to the absent of information for several criteria. Therefore, this opens another question, the aggregation procedure, which as it is known from the specialised literature on \czm{DM}, there are many proposals and also with variability of results. Additionally, from a global evaluation of a review as an evaluation, we can also establish potential procedures for filtering them as neutral opinion following the mentioned study, detecting and filtering neutrality in sentiment analysis \cite{valdivia2018}.}

\subsection{Sensitivity}
\ph{The conducted experiments in this section lead us to realize that ChatGPT presents a high level of sensitivity in the responses, as a level of uncertainty previously mentioned and study in \cite{huang2023look}}. The \new{evaluations} extracted from the reviews varies when making different requests. It becomes particularly evident when the reviews are complex and reflect a mix of opinions. For example, \czm{Figures \ref{review_1}, \ref{review_2}, and \ref{review_3} present} reviews whose sentiments extracted by ChatGPT differ according to different executions, showing that in most experiments ChatGPT identifies a negative opinion and in others neutral. Subsequently, we analyze them.

\begin{figure}[ht]
\examplebox{Review 1}
{Getting into the Ivy ,back in the day ,was a real achievement. Constant phone calls to be made and if you hadn’t been before ,and weren’t on their database, it could be a frustrating process. The first time we went ,I was bowled over. The food was fairly simple but absolutely delicious and the atmosphere was terrific - a real occasion. You were almost guaranteed to see some sort of minor celebrity almost anytime. The staff were the stars though. The matre d’ (name escapes me) and other staff were real charmers. No matter how much you were disappointed in the table allocated (and we dined in what was politely called `social Siberia’ several times!), they could always make it feel better.We revisited earlier this month one Saturday lunchtime. No problem getting a table - booked on the internet a couple of weeks beforehand. Arrived and were seated in the prime location - unbelievable. Then commenced a pretty ordinary two courses served by staff who were ok but just ok. No charm, no schmooze! Food was perfectly edible but not memorable. My wife pronounced her fish pie `bland’ and my pork belly wasn’t in the same league as the best I’ve had. Wine was goosd though and reasonable.No celebrities - not even small ones. Lots of ladies from Essex and people taking photographs of themselves at the Ivy.I still enjoyed it and if I hadn’t been before, I might be giving it a different rating but when the bill came, our two courses and drinks were almost as much as the set lunch at Murano or Maze or several other far superior establishments.If you're in the area, give it a try for old times sake. If you want buzzy - go to the Wolseley instead. If you want great food, check out any one of London's other many fine restaurants.}
\caption{\czm{Review from TripR-2020Large review in which ChatGPT is sensitive when inferring the main \new{evaluation}.}}
\label{review_1}
\end{figure}

\begin{figure}[ht]
\examplebox{Review 2}
{The food is ok, nothing to write home about, Skylon has an equally good view and is better and better value}
\caption{\czm{Review from TripR-2020Large in which ChatGPT is sensitive when inferring the main \new{evaluation}. }}
\label{review_2}
\end{figure}

\begin{figure}[ht]
\examplebox{Review 3}
{Sheekey was given a new lease of life some years ago when it was acquired by the owner of The Ivy.You have a choice here between the Restaurant and the Oyster Bar.Unsurprisingly, the Restaurant offers the more extensive choice of dishes whilst the Oyster Bar menu contains a smaller selection of not simply oysters but also a good range of seafood dishes. The Bar is my favourite, first, as a result of the really friendly and helpful staff and ,secondly, the informal layout of the bar and tables which always seem to generate a really good buzz. The bar is also particularly useful if you are eating alone.Not cheap but quality rarely is.}
\caption{\czm{Review from TripR-2020Large in which ChatGPT is not sensitive when inferring the main \new{evaluation}.}}
\label{review_3}
\end{figure}

\begin{figure}[!ht]
    \centering
    \includegraphics[width=0.9\linewidth]{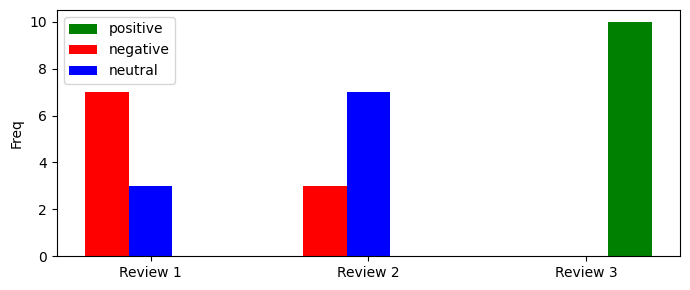}
    \caption{Sensitivity in the scenario with primary polarities.}
    \label{fig:sensi-sce1}
\end{figure}

In the scenario \ref{cs_3labels}, we get Review 1, 2, and 3 have associated \textit{negative}, \textit{neutral}, and \textit{positive} polarities, respectively. Now, we made ten requests for each review in order to conduct the sensitivity analysis. Results are shown in Figure \ref{fig:sensi-sce1}. We notice: 
\begin{enumerate*}[label=(\arabic*)] 
\item the sentiments expressed in reviews 1 and 2 varies from \textit{negative} to \textit{neutral} as texts are confusing, and
\item review 3 gets \textit{positive} polarity in all proofs as it is quite immediate. 
\end{enumerate*} 

We perform a similar analysis with the above three reviews inferring the evaluations as numerical scores in the scenario \ref{cs_num}. Table \ref{tab:sensib} presents the ChatGPT outputs of ten independent requests for each review. We realize that the numerical scores have significant variation among different proofs, so this study reveals that it would be appropriate to handle intervals. 

\begin{table}[!ht]
    \centering
    \begin{tabular}{c|ccc}
         \czm{Experiment} &  Review 1 & Review 2 & Review 3 \\
         \hline
       1 &  0.4 & 0.3 & 0.8 \\
       2 & 0.32 & 0.2 & 0.8 \\
       3 & 0.325 & 0.3 & 0.85 \\
       4 & 0.35 & 0.4 & 0.9 \\
       5 & 0.4 & 0.55 & 0.8 \\
       6 & 0.4 & 0.3 & 0.85 \\
       7 & 0.44 & 0.35 & 0.85 \\
       8 & 0.4 & 0.2 & 0.85 \\
       9 & 0.425 & 0.25 & 0.9 \\
       10 & 0.45 & 0.3 & 0.75 \\
    \end{tabular}
    \caption{Sensitivity in the scenario with numerical scores.}
    \label{tab:sensib}
\end{table}

Finally, we analyze the sensitivity of the general \new{evaluation} provided by ChatGPT as a CDM system (see Section \ref{cs_endtoend}). This analysis is performed on all the reviews for each restaurant by requesting a numerical score for each one. Table \ref{tab:general-sens} shows the results obtained in five requests. The NaN values indicate that it does not provide a suitable response as a numerical score. We observe that experiments 1 and 2 show a different order between restaurants 3 and 4, getting similar rankings to those obtained in the first three scenarios. 
\begin{table}[!ht]
    \centering
    \begin{tabular}{c|cccc}
         \czm{Experiment} &  The Oxo Tower & The Wolseley & The Ivy & J. Sheekey \\
         \hline
        1 & 0.59 & 0.75 & 0.8 & 0.75 \\
        2 & 0.6 & NaN & 0.7 & 0.759 \\
        3 & 0.7 & 0.7 & NaN & 0.7 \\
        4 & 0.617 & 0.7 & NaN & 0.7 \\
        5 & 0.676 & 0.8 & NaN & 0.7 \\
    \end{tabular}
    \caption{Sensitivity of ChatGPT as a DM tool.}
    \label{tab:general-sens}
\end{table}

This behaviour continues to show the inherent uncertainty that exists in extracting a simple \new{evaluation} from a review that can be complex in the opinions that reflect different nuances and aspects to be considered. It is an open issue to design ChatGPT based robust \czm{DM} models dealing with the sensitivity together with the inconsistencies. One possibility might be to make several queries and to keep the modal value if linguistic, or the average \new{value} if numerical. 

\subsection{Explainability}
\ph{Explainability is an important artificial intelligence (AI) challenge. In \cite{arrieta2020explainable} is defined the explainable AI (XAI) as follows:
\it{"Given an audience, an explainable AI is one that produces details or reasons to make its functioning clear or easy to understand."}. A fundamental XAI element is the explanations associated to any AI systems distilling decision \cite{bodria2023benchmarking, longo2024explainable}. Likewise, in the design of computational models in complex scenarios such as the one we are in, as discussed in \cite{zuheros2023decision}, the ability to interpret and justify the decisions, reasoning or significant features of the model considered to make the predictions is required.}

ChatGPT is a black-box artificial intelligence model that does not inherently offer detailed explanations of its responses. It is unable to apply most explainability methods because, among other reasons, it does not provide access to external tools, such as to apply feature attribution methods like LIME, or to visual interfaces, such as to apply attention visualization. 

We can set up a first explanation scenario with ChatGPT by asking a direct question about it, and therefore get insights into its \czm{DM} process. The answer for the following example, as we can observe, is more complex than the simple evaluation.

\begin{example}
The following review is analyzed by ChatGPT. \czm{We consider} the scenario with primary polarities (Section \ref{cs_3labels}). Subsequently, we update the corresponding prompt by adding at the end: ``Then, explain your response.''. This way, ChatGPT indicates a \textit{negative} \new{evaluation} and offers an explanation of the response as shown below. The explanation is quite complex and, in general, we notice a tendency to focus on keywords or phrases that have strong emotional connotations.
\examplebox{Review to infer primary polarity}
{Very crowded and overbooking of tables. Impossible to get a river view table. Food reasonable but expensive. Trades on its good name}
\examplebox{ChatGPT response with explanation}
{Negative

Explanation:
The text includes negative elements such as ``very crowded" and ``overbooking of tables," indicating a frustrating and inconvenient experience. Additionally, the statement ``Impossible to get a river view table" suggests dissatisfaction with the seating arrangement. Although the food is described as ``reasonable," the negative aspects mentioned earlier overshadow it. The phrase ``Food reasonable but expensive" indicates that the prices may not be justified by the quality of the experience. The last part of the sentence, ``Trades on its good name," suggests that the establishment may be relying on its reputation rather than delivering an exceptional experience. Overall, the text expresses a negative sentiment due to the presence of several negative aspects and criticisms.}
\end{example}

\ph{The explainability for DM based on LLMs is an important challenge that requires deep studies to provide explanations on the decisions to increase user trust in these models. It also would be of great interest 
to identify the most relevant aspects to distil the evaluation \new{and to explain them}. }   


\section{\ph{Conclusions and future work on challenges}}\label{sec5}
This paper consider\czm{s} the use of LLMs, and in particular the use of ChatGPT, to design \czm{DM} models to process natural language \ph{reviews} from social media, \new{able} to distil \ph{evaluations as} the wisdom of crowd which is embedded in the multitude of \czm{texts}. The \czm{five proposed} scenarios distilling primary \textit{positive}, \textit{negative}, and \textit{neutral} polarities, numerical scores or linguistic term set scores, show the great potential of LLMs \czm{using prompt design strategies for designing CDM. This opens a new scenario in DM based on \new{evaluations from the opinions of expert reviews.}}

The main findings and conclusions of this study are:
\begin{itemize}
    \item ChatGPT prompts must be concise, clearly specifying the output format. 
    \item The flexibility of ChatGPT allows to extract the evaluations from \ph{opinions as decisions} both as numerical and linguistic assessments, and also from a category ontology for the alternatives associated to criteria. 
    \item ChatGPT presents challenges to address in the opinions analysis for CDM. We have discussed three important \czm{open issues} such as the inconsistencies processing opinions, sensitivity and explainabilty.
\end{itemize}

In short, the present study  shows the potential possibilities, at the beginning, of using  LLMs in the design of \czm{DM} models that process expert's reviews expresses in natural language. As future work on challenges, a wide range of possibilities opens up. The following paths emerge from the studies carried out: 
\ph{\begin{itemize}
\item \it{Discussed challenges.} The open issues discussed above, consistency, sensitivity and explainability, require the design of more robust and explainable \czm{DM} models. They must be tackled with attention for the design of new CDM models based on LLMs. 
\item \it{The prompt engineering.} It is worth thinking about exploring the use of LLMs in general and  ChatGPT in particular with enriched and more complex queries in prompts, which may allow the design of different \czm{DM} models that work next to common sense in human \czm{DM}. 
\item \it{Handling literary \new{tweets} and turns. } Other open issue is the ability of DM models to address some NLP challenges of interest for sentiment extraction, such as sarcasm, irony and other literary turns of phrase present in the language.
\item \it{CDM based on specific LLMs.} It is worth asking about other LLMs, and 
 to what extent a specific LLM model can be designed for DM in a  particular domain, including enabling  model learning, as mentioned in \cite{yang2023autogpt}, presenting a benchmark study of Auto-GPT, an autonomous agent inspired by ChatGPT. 
 \item \it{ LLMs safety.} At last but not least is the LLMs safety, due to they exhibit inherent issues, vulnerabilities, and unintended bugs. These can impact their reliability and safety. Among others, problems such us certifying LLMs safety against adversarial prompting for DM models, how does LLMs safety training fail, and towards improved safety alignment of LLMs via a human-preference dataset, (see recent literature \cite{kumar2023certifying, wei2024jailbroken, ji2024beavertails}), require to be tackled. 
\end{itemize}}

\ph{ As a final reflection, we are facing an emerging and fascinating new scenario with the presence of LLMs that allows the assessment of opinions and the design of DM models. This is further enhanced by the fact that in the era of social media platforms, opinions are naturally expressed through written reviews.
Advances in LLM development will enable the design of new DM models that accurately attack all three mentioned pathways on the basis of prompts engineering, and will enable progress on DM models that emulate common-sense human reasoning in the opinion crowd process.}

\section*{Acknowledgments}
F. Herrera and R. Montes acknowledge funding support by Grant “Convenio de Colaboración entre la Universidad de Granada y la S.M.E Instituto Nacional de Ciberseguridad de España M.P., S.A. para la Promoción de Proyectos Estratégicos de Ciberseguridad en España”, funded by S.M.E. Instituto Nacional de Ciberseguridad de España M.P. S.A. (“INCIBE”) and European Union – NextGenerationEU.

\newpage

\bibliography{main}

\appendix
\section{Detailed results for the scenario multi-criteria}
\label{appendix:tables_MC}
Tables \ref{tab:freq-cat-0}, \ref{tab:freq-cat-1}, \ref{tab:freq-cat-2}, and \ref{tab:freq-cat-3} present the statistics associated to the restaurants in the scenario \ref{sce_mc}. The columns \textit{\#Positive}, \textit{\#Neutral}, and \textit{\#Negative} presents the number of reviews with positive, negative and neutral opinions. The \textit{\#None} column indicates the number of reviews without opinions according to ChatGPT. The \textit{mean} column is calculated by asking ChatGPT for a numerical result for the categories and averaging them. The \textit{Positive\%} column is computed by the percentage of the positive relative frequency ignoring the \textit{None} values.
\begin{table*}[!hb]
    \centering
    \begin{tabular}{ccccccc}
        \multirow{2}{*}{Category}&\multicolumn{4}{c}{Sentiment}&\multirow{2}{*}{Mean}&\multirow{2}{*}{Positive\%}\\
        &\#Positive&\#Neutral&\#Negative&\#None & &\\
        \hline
        FOOD\#QUALITY&60&8&12&15&0.78&75.00\\
        FOOD\#STYLE\_OPTIONS&13&5&1&76&0.84&68.42\\
        FOOD\#PRICES&7&11&15&62&0.56&21.21\\
        DRINKS\#QUALITY&41&3&0&51&0.90&93.18\\
        DRINKS\#PRICES&4&11&7&73&0.65&18.18\\
        RESTAURANT\#GENERAL&71&7&9&8&0.90&81.61\\
        LOCATION\#GENERAL&56&3&0&36&0.90&94.92\\
        SERVICE\#GENERAL&67&3&11&14&0.85&82.72\\
        AMBIENCE\#GENERAL&73&3&7&12&0.83&87.95\\
        RESTAURANT\#PRICES&9&7&16&63&0.64&28.12\\
        DRINKS\#STYLE\_OPTIONS&9&3&0&83&0.88&75.00\\
        RESTAURANT\#MISCELLANEOUS&10&0&1&84&1.00&90.91\\
        \hline
        Mean& & & & & 0.81 & 68.10 \\
    \end{tabular}
    \caption{Sentiment frequency associated with the ontology categories for The Oxo Tower restaurant.}
    \label{tab:freq-cat-0}
\end{table*}

\begin{table*}[!hb]
    \centering
    \begin{tabular}{ccccccc}
        \multirow{2}{*}{Category}&\multicolumn{4}{c}{Sentiment}&\multirow{2}{*}{Mean}&\multirow{2}{*}{Positive\%}\\
        &\#Positive&\#Neutral&\#Negative&\#None & &\\
        \hline
        FOOD\#QUALITY&104&8&11&11&0.84&84.55\\
        FOOD\#STYLE\_OPTIONS&26&13&0&95&0.83&66.67\\
        FOOD\#PRICES&17&9&18&90&0.64&38.64\\
        DRINKS\#QUALITY&29&11&4&90&0.75&65.91\\
        DRINKS\#PRICES&6&8&7&113&0.50&28.57\\
        RESTAURANT\#GENERAL&109&5&16&4&0.87&83.85\\
        LOCATION\#GENERAL&32&2&0&100&0.85&94.12\\
        SERVICE\#GENERAL&92&8&18&16&0.82&77.97\\
        AMBIENCE\#GENERAL&94&3&12&25&0.87&86.24\\
        RESTAURANT\#PRICES&18&11&17&88&0.60&39.13\\
        DRINKS\#STYLE\_OPTIONS&4&10&0&120&0.80&28.57\\
        RESTAURANT\#MISCELLANEOUS&16&1&1&116&nan&88.89\\
        \hline
                Mean& & & & & 0.76 & 65.26 \\
    \end{tabular}
    \caption{Sentiment frequency associated with the ontology categories for The Wolseley restaurant.}
    \label{tab:freq-cat-1}
\end{table*}

\begin{table*}[!hb]
    \centering
    \begin{tabular}{ccccccc}
        \multirow{2}{*}{Category}&\multicolumn{4}{c}{Sentiment}&\multirow{2}{*}{Mean}&\multirow{2}{*}{Positive\%}\\
        &\#Positive&\#Neutral&\#Negative&\#None & &\\
        \hline
        FOOD\#QUALITY&123&6&14&5&0.86&86.01\\
        FOOD\#STYLE\_OPTIONS&29&16&1&102&0.86&63.04\\
        FOOD\#PRICES&19&16&19&94&0.67&35.19\\
        DRINKS\#QUALITY&33&13&0&102&0.88&71.74\\
        DRINKS\#PRICES&4&12&7&125&0.60&17.39\\
        RESTAURANT\#GENERAL&124&10&7&7&0.83&87.94\\
        LOCATION\#GENERAL&18&7&0&123&1.00&72.00\\
        SERVICE\#GENERAL&115&6&16&11&0.86&83.94\\
        AMBIENCE\#GENERAL&101&8&8&31&0.88&86.32\\
        RESTAURANT\#PRICES&18&16&19&95&0.58&33.96\\
        DRINKS\#STYLE\_OPTIONS&7&11&0&130&0.00&38.89\\
        RESTAURANT\#MISCELLANEOUS&21&7&2&118&0.95&70.00\\
        \hline
                Mean& & & & & 0.75 & 62.20 \\

    \end{tabular}
    \caption{Sentiment frequency associated with the ontology categories for The Ivy restaurant.}
    \label{tab:freq-cat-2}
\end{table*}

\begin{table*}[!hb]
    \centering
    \begin{tabular}{ccccccc}
        \multirow{2}{*}{Category}&\multicolumn{4}{c}{Sentiment}&\multirow{2}{*}{Mean}&\multirow{2}{*}{Positive\%}\\
        &\#Positive&\#Neutral&\#Negative&\#None & &\\
        \hline
        FOOD\#QUALITY&76&1&7&13&0.88&90.48\\
        FOOD\#STYLE\_OPTIONS&15&3&0&79&0.83&83.33\\
        FOOD\#PRICES&10&8&14&65&0.62&31.25\\
        DRINKS\#QUALITY&29&2&0&66&0.87&93.55\\
        DRINKS\#PRICES&5&5&2&85&0.66&41.67\\
        RESTAURANT\#GENERAL&77&5&9&6&0.87&84.62\\
        LOCATION\#GENERAL&20&3&1&73&0.93&83.33\\
        SERVICE\#GENERAL&73&2&8&14&0.86&87.95\\
        AMBIENCE\#GENERAL&63&4&7&23&0.81&85.14\\
        RESTAURANT\#PRICES&12&7&16&62&0.61&34.29\\
        DRINKS\#STYLE\_OPTIONS&4&4&0&89&0.70&50.00\\
        RESTAURANT\#MISCELLANEOUS&9&0&2&86&nan&81.82\\
        \hline
                Mean& & & & & 0.78 & 70.62 \\

    \end{tabular}
    \caption{Sentiment frequency associated with the ontology categories for J. Sheekey restaurant.}
    \label{tab:freq-cat-3}
\end{table*}

\end{document}

%% file: ontology.tex
\begin{figure}[!hbt]
\centering
\tikzset{every picture/.style={line width=0.75pt}} 
\resizebox{0.9\textwidth}{5cm}{  
\begin{tikzpicture}[
  root/.style={rectangle,draw,fill=blue!15},
  criteria/.style={rectangle,draw,fill=blue!10},
  attrib/.style={rectangle,draw,fill=orange!10,rounded corners=.8ex},
  grandchild/.style={grow=down,xshift=1em,anchor=west,
    edge from parent path={(\tikzparentnode.south) |- (\tikzchildnode.west)}},
  first/.style={level distance=6ex},
  second/.style={level distance=12ex},
  third/.style={level distance=18ex},
  level 1/.style={sibling distance=5em}]
    \coordinate
      node[root,anchor=center]{Restaurant}
    child{node[criteria] {restaurant}
      child[grandchild,first] {node[attrib]{general}}
      child[grandchild,second] {node[attrib]{prices}}
      child[grandchild,third] {node[attrib] {miscellaneous}}}
    child {node[criteria] {service}
      child[grandchild,first] {node[attrib]{general}}}
    child{node[criteria] {food}
      child[grandchild,first] {node[attrib]{style}}
      child[grandchild,second] {node[attrib]{quality}}
      child[grandchild,third] {node[attrib] {prices}}}
    child {node[criteria] {ambience}
      child[grandchild,first] {node[attrib]{general}}}
    child {node[criteria]{drinks}
      child[grandchild,first] {node[attrib]{style}}
      child[grandchild,second] {node[attrib]{quality}}
      child[grandchild,third] {node[attrib] {prices}}}
    child {node[criteria] {location}
      child[grandchild,first] {node[attrib]{general}}};
\end{tikzpicture}
}

\caption{Category ontology for the alternatives.}
\label{fig_ontology}
\end{figure}

%% file: main.bbl
\begin{thebibliography}{43}
\expandafter\ifx\csname natexlab\endcsname\relax\def\natexlab#1{#1}\fi
\providecommand{\url}[1]{\texttt{#1}}
\providecommand{\href}[2]{#2}
\providecommand{\path}[1]{#1}
\providecommand{\DOIprefix}{doi:}
\providecommand{\ArXivprefix}{arXiv:}
\providecommand{\URLprefix}{URL: }
\providecommand{\Pubmedprefix}{pmid:}
\providecommand{\doi}[1]{\href{http://dx.doi.org/#1}{\path{#1}}}
\providecommand{\Pubmed}[1]{\href{pmid:#1}{\path{#1}}}
\providecommand{\bibinfo}[2]{#2}
\ifx\xfnm\relax \def\xfnm[#1]{\unskip,\space#1}\fi
\bibitem[{Liu(2012)}]{liu2012sentiment}
\bibinfo{author}{B.~Liu},
\newblock \bibinfo{title}{Sentiment analysis and opinion mining},
\newblock \bibinfo{journal}{Synthesis lectures on human language technologies} \bibinfo{volume}{5} (\bibinfo{year}{2012}) \bibinfo{pages}{1--167}.
\bibitem[{Trillo et~al.(2023)Trillo, Herrera-Viedma, Morente-Molinera, and Cabrerizo}]{trillo2023group}
\bibinfo{author}{J.~R. Trillo}, \bibinfo{author}{E.~Herrera-Viedma}, \bibinfo{author}{J.~A. Morente-Molinera}, \bibinfo{author}{F.~J. Cabrerizo},
\newblock \bibinfo{title}{A group decision-making method based on the experts’ behavior during the debate},
\newblock \bibinfo{journal}{IEEE Transactions on Systems, Man, and Cybernetics: Systems} \bibinfo{volume}{53} (\bibinfo{year}{2023}) \bibinfo{pages}{5796--5808}.
\bibitem[{Herrera-Viedma et~al.(2021)}]{viedma2021decision}
\bibinfo{author}{E.~Herrera-Viedma}, et~al.,
\newblock \bibinfo{title}{{Revisiting Fuzzy and Linguistic Decision Making: Scenarios and Challenges for Making Wiser Decisions in a Better Way}},
\newblock \bibinfo{journal}{IEEE Transactions on Systems, Man, and Cybernetics: Systems} \bibinfo{volume}{51} (\bibinfo{year}{2021}) \bibinfo{pages}{191--208}.
\bibitem[{Li et~al.(2022)Li, Dong, Pedrycz, and Herrera}]{9253574}
\bibinfo{author}{C.-C. Li}, \bibinfo{author}{Y.~Dong}, \bibinfo{author}{W.~Pedrycz}, \bibinfo{author}{F.~Herrera},
\newblock \bibinfo{title}{{Integrating Continual Personalized Individual Semantics Learning in Consensus Reaching in Linguistic Group Decision Making}},
\newblock \bibinfo{journal}{IEEE Transactions on Systems, Man, and Cybernetics: Systems} \bibinfo{volume}{52} (\bibinfo{year}{2022}) \bibinfo{pages}{1525--1536}.
\bibitem[{Li et~al.(2023)}]{mengqi2023social}
\bibinfo{author}{M.~Li}, et~al.,
\newblock \bibinfo{title}{{A Trust Risk Dynamic Management Mechanism Based on Third-Party Monitoring for the Conflict-Eliminating Process of Social Network Group Decision Making}},
\newblock \bibinfo{journal}{IEEE Transactions on Cybernetics} \bibinfo{volume}{53} (\bibinfo{year}{2023}) \bibinfo{pages}{3399--3413}.
\bibitem[{Liu et~al.(2022)Liu, Xu, Gong, and Herrera}]{LIU2022220}
\bibinfo{author}{X.~Liu}, \bibinfo{author}{Y.~Xu}, \bibinfo{author}{Z.~Gong}, \bibinfo{author}{F.~Herrera},
\newblock \bibinfo{title}{Democratic consensus reaching process for multi-person multi-criteria large scale decision making considering participants’ individual attributes and concerns},
\newblock \bibinfo{journal}{Information Fusion} \bibinfo{volume}{77} (\bibinfo{year}{2022}) \bibinfo{pages}{220--232}.
\bibitem[{Zuheros et~al.(2022)Zuheros, Mart{\'\i}nez-C{\'a}mara, Herrera-Viedma, and Herrera}]{zuheros2022crowd}
\bibinfo{author}{C.~Zuheros}, \bibinfo{author}{E.~Mart{\'\i}nez-C{\'a}mara}, \bibinfo{author}{E.~Herrera-Viedma}, \bibinfo{author}{F.~Herrera},
\newblock \bibinfo{title}{{Crowd Decision Making: Sparse Representation Guided by Sentiment Analysis for Leveraging the Wisdom of the Crowd}},
\newblock \bibinfo{journal}{IEEE Transactions on Systems, Man, and Cybernetics: Systems} \bibinfo{volume}{53} (\bibinfo{year}{2022}) \bibinfo{pages}{369--379}.
\bibitem[{Surowiecki(2005)}]{surowiecki2005wisdom}
\bibinfo{author}{J.~Surowiecki}, \bibinfo{title}{The wisdom of crowds}, \bibinfo{publisher}{Anchor}, \bibinfo{year}{2005}.
\bibitem[{Pang et~al.(2008)Pang, Lee et~al.}]{pang2008opinion}
\bibinfo{author}{B.~Pang}, \bibinfo{author}{L.~Lee}, et~al.,
\newblock \bibinfo{title}{Opinion mining and sentiment analysis},
\newblock \bibinfo{journal}{{Foundations and Trends® in Information Retrieval}} \bibinfo{volume}{2} (\bibinfo{year}{2008}) \bibinfo{pages}{1--135}.
\bibitem[{Ren and Wu(2018)}]{ren2018innovative}
\bibinfo{author}{R.~Ren}, \bibinfo{author}{D.~Wu},
\newblock \bibinfo{title}{An innovative sentiment analysis to measure herd behavior},
\newblock \bibinfo{journal}{IEEE Transactions on Systems, Man, and Cybernetics: Systems} \bibinfo{volume}{50} (\bibinfo{year}{2018}) \bibinfo{pages}{3841--3851}.
\bibitem[{Wen et~al.(2019)Wen, Wei, Yang, Guo, Zeng, Huang, and Chen}]{wen2019memristive}
\bibinfo{author}{S.~Wen}, \bibinfo{author}{H.~Wei}, \bibinfo{author}{Y.~Yang}, \bibinfo{author}{Z.~Guo}, \bibinfo{author}{Z.~Zeng}, \bibinfo{author}{T.~Huang}, \bibinfo{author}{Y.~Chen},
\newblock \bibinfo{title}{{Memristive LSTM network for sentiment analysis}},
\newblock \bibinfo{journal}{IEEE Transactions on Systems, Man, and Cybernetics: Systems} \bibinfo{volume}{51} (\bibinfo{year}{2019}) \bibinfo{pages}{1794--1804}.
\bibitem[{Radford et~al.(2019)}]{radford2019language}
\bibinfo{author}{A.~Radford}, et~al.,
\newblock \bibinfo{title}{Language models are unsupervised multitask learners},
\newblock \bibinfo{journal}{OpenAI blog} \bibinfo{volume}{1} (\bibinfo{year}{2019}) \bibinfo{pages}{9}.
\bibitem[{Brants et~al.(2007)}]{brants2007large}
\bibinfo{author}{T.~Brants}, et~al.,
\newblock \bibinfo{title}{Large language models in machine translation},
\newblock \bibinfo{journal}{Proceedings of the 2007 Joint Conference on Empirical Methods in Natural Language Processing and Computational Natural Language Learning}  (\bibinfo{year}{2007}).
\bibitem[{{OpenAI}(2023)}]{openai2021chatgpt}
\bibinfo{author}{{OpenAI}}, \bibinfo{title}{{ChatGPT}}, \bibinfo{howpublished}{\url{https://openai.com/chatgpt}}, \bibinfo{year}{2023}.
\bibitem[{Triantaphyllou and Triantaphyllou(2000)}]{triantaphyllou2000multi}
\bibinfo{author}{E.~Triantaphyllou}, \bibinfo{author}{E.~Triantaphyllou}, \bibinfo{title}{Multi-criteria decision making methods}, \bibinfo{publisher}{Springer}, \bibinfo{year}{2000}.
\bibitem[{Losee(2001)}]{losee2001natural}
\bibinfo{author}{R.~M. Losee},
\newblock \bibinfo{title}{Natural language processing in support of decision-making: phrases and part-of-speech tagging},
\newblock \bibinfo{journal}{Information processing \& management} \bibinfo{volume}{37} (\bibinfo{year}{2001}) \bibinfo{pages}{769--787}.
\bibitem[{Zuheros et~al.(2021)Zuheros, Mart{\'\i}nez-C{\'a}mara, Herrera-Viedma, and Herrera}]{zuheros2021sentiment}
\bibinfo{author}{C.~Zuheros}, \bibinfo{author}{E.~Mart{\'\i}nez-C{\'a}mara}, \bibinfo{author}{E.~Herrera-Viedma}, \bibinfo{author}{F.~Herrera},
\newblock \bibinfo{title}{{Sentiment analysis based multi-person multi-criteria decision making methodology using natural language processing and deep learning for smarter decision aid. Case study of restaurant choice using TripAdvisor reviews}},
\newblock \bibinfo{journal}{Information Fusion} \bibinfo{volume}{68} (\bibinfo{year}{2021}) \bibinfo{pages}{22--36}.
\bibitem[{Cao et~al.(2022)Cao, Xu, Yin, and Pan}]{cao2022risky}
\bibinfo{author}{J.~Cao}, \bibinfo{author}{X.~Xu}, \bibinfo{author}{X.~Yin}, \bibinfo{author}{B.~Pan},
\newblock \bibinfo{title}{A risky large group emergency decision-making method based on topic sentiment analysis},
\newblock \bibinfo{journal}{Expert Systems with Applications} \bibinfo{volume}{195} (\bibinfo{year}{2022}) \bibinfo{pages}{116527}.
\bibitem[{Tayal et~al.(2023)Tayal, Yadav, and Arora}]{tayal2023personalized}
\bibinfo{author}{D.~K. Tayal}, \bibinfo{author}{S.~K. Yadav}, \bibinfo{author}{D.~Arora},
\newblock \bibinfo{title}{{Personalized ranking of products using aspect-based sentiment analysis and Plithogenic sets}},
\newblock \bibinfo{journal}{Multimedia Tools and Applications} \bibinfo{volume}{82} (\bibinfo{year}{2023}) \bibinfo{pages}{1261--1287}.
\bibitem[{Zhu et~al.(2023)Zhu, Xu, and Pan}]{zhu2023method}
\bibinfo{author}{Y.~Zhu}, \bibinfo{author}{X.~Xu}, \bibinfo{author}{B.~Pan},
\newblock \bibinfo{title}{A method for the dynamic collaboration of the public and experts in large-scale group emergency decision-making: Using social media data to evaluate the decision-making quality},
\newblock \bibinfo{journal}{Computers \& Industrial Engineering} \bibinfo{volume}{176} (\bibinfo{year}{2023}) \bibinfo{pages}{108943}.
\bibitem[{Hossain et~al.(2023)}]{hossain2023natural}
\bibinfo{author}{E.~Hossain}, et~al.,
\newblock \bibinfo{title}{{Natural language processing in electronic health records in relation to healthcare decision-making: A systematic review}},
\newblock \bibinfo{journal}{Comput. Biol. Med} \bibinfo{volume}{155} (\bibinfo{year}{2023}) \bibinfo{pages}{106649}.
\bibitem[{Valdivia et~al.(2017)Valdivia, Luzón, and Herrera}]{valdivia2017}
\bibinfo{author}{A.~Valdivia}, \bibinfo{author}{M.~V. Luzón}, \bibinfo{author}{F.~Herrera},
\newblock \bibinfo{title}{{Sentiment analysis in TripAdvisor}},
\newblock \bibinfo{journal}{IEEE Intelligent Systems} \bibinfo{volume}{32} (\bibinfo{year}{2017}) \bibinfo{pages}{72--77}.
\bibitem[{Zhao et~al.(2023)Zhao, Jin, Del~Ser, and Yang}]{zhao2023chatagri}
\bibinfo{author}{B.~Zhao}, \bibinfo{author}{W.~Jin}, \bibinfo{author}{J.~Del~Ser}, \bibinfo{author}{G.~Yang},
\newblock \bibinfo{title}{{ChatAgri: Exploring Potentials of ChatGPT on Cross-linguistic Agricultural Text Classification}},
\newblock \bibinfo{journal}{arXiv:2305.15024}  (\bibinfo{year}{2023}).
\bibitem[{Koco{\'n} et~al.(2023)}]{kocon2023chatgpt}
\bibinfo{author}{J.~Koco{\'n}}, et~al.,
\newblock \bibinfo{title}{{ChatGPT: Jack of all trades, master of none}},
\newblock \bibinfo{journal}{Information Fusion} \bibinfo{volume}{99} (\bibinfo{year}{2023}) \bibinfo{pages}{101861}.
\bibitem[{Maddigan and Susnjak(2023)}]{maddigan2023chat2vis}
\bibinfo{author}{P.~Maddigan}, \bibinfo{author}{T.~Susnjak},
\newblock \bibinfo{title}{{Chat2vis: Generating data visualisations via natural language using ChatGPT, Codex and GPT-3 Large Language Models}},
\newblock \bibinfo{journal}{arXiv:2302.02094}  (\bibinfo{year}{2023}).
\bibitem[{Sorin et~al.(2023)}]{sorin2023large}
\bibinfo{author}{V.~Sorin}, et~al.,
\newblock \bibinfo{title}{{Large language model (ChatGPT) as a support tool for breast tumor board}},
\newblock \bibinfo{journal}{NPJ Breast Cancer} \bibinfo{volume}{9} (\bibinfo{year}{2023}) \bibinfo{pages}{44}.
\bibitem[{Amin et~al.(2023)Amin, Cambria, and Schuller}]{Ami2023}
\bibinfo{author}{M.~M. Amin}, \bibinfo{author}{E.~Cambria}, \bibinfo{author}{B.~W. Schuller},
\newblock \bibinfo{title}{{Can ChatGPT's Responses Boost Traditional Natural Language Processing?}},
\newblock \bibinfo{journal}{arXiv:2307.04648}  (\bibinfo{year}{2023}).
\bibitem[{Yang et~al.(2023)Yang, Yue, and He}]{yang2023autogpt}
\bibinfo{author}{H.~Yang}, \bibinfo{author}{S.~Yue}, \bibinfo{author}{Y.~He},
\newblock \bibinfo{title}{{Auto-GPT for Online Decision Making: Benchmarks and Additional Opinions}},
\newblock \bibinfo{journal}{arXiv:2306.02224}  (\bibinfo{year}{2023}).
\bibitem[{Wang et~al.(2023)}]{wang2023chatgpt}
\bibinfo{author}{F.-Y. Wang}, et~al.,
\newblock \bibinfo{title}{{ChatGPT for computational social systems: From conversational applications to human-oriented operating systems}},
\newblock \bibinfo{journal}{IEEE Transactions on Computational Social Systems} \bibinfo{volume}{10} (\bibinfo{year}{2023}) \bibinfo{pages}{414--425}.
\bibitem[{Gao et~al.(2023)}]{gao2023chat}
\bibinfo{author}{Y.~Gao}, et~al.,
\newblock \bibinfo{title}{{Chat With ChatGPT on Interactive Engines for Intelligent Driving}},
\newblock \bibinfo{journal}{IEEE Transactions on Intelligent Vehicles} \bibinfo{volume}{8} (\bibinfo{year}{2023}) \bibinfo{pages}{2034--2036}.
\bibitem[{Guo et~al.(2023)Guo, Lu, Dou, and Wang}]{guo2023can}
\bibinfo{author}{C.~Guo}, \bibinfo{author}{Y.~Lu}, \bibinfo{author}{Y.~Dou}, \bibinfo{author}{F.-Y. Wang},
\newblock \bibinfo{title}{{Can ChatGPT boost artistic creation: The need of imaginative intelligence for parallel art}},
\newblock \bibinfo{journal}{IEEE/CAA Journal of Automatica Sinica} \bibinfo{volume}{10} (\bibinfo{year}{2023}) \bibinfo{pages}{835--838}.
\bibitem[{Nikolic et~al.(2023)}]{nikolic2023chatgpt}
\bibinfo{author}{S.~Nikolic}, et~al.,
\newblock \bibinfo{title}{{ChatGPT versus engineering education assessment: a multidisciplinary and multi-institutional benchmarking and analysis of this generative artificial intelligence tool to investigate assessment integrity}},
\newblock \bibinfo{journal}{{European Journal of Engineering Education}} \bibinfo{volume}{48} (\bibinfo{year}{2023}) \bibinfo{pages}{559--614}.
\bibitem[{White and et~al.(2023)}]{white2023prompt}
\bibinfo{author}{J.~White}, \bibinfo{author}{et~al.},
\newblock \bibinfo{title}{{A Prompt Pattern Catalog to Enhance Prompt Engineering with ChatGPT}},
\newblock \bibinfo{journal}{arXiv:2302.11382}  (\bibinfo{year}{2023}).
\bibitem[{Mart{\'\i}nez et~al.(2015)Mart{\'\i}nez, Rodriguez, and Herrera}]{martinez20152}
\bibinfo{author}{L.~Mart{\'\i}nez}, \bibinfo{author}{R.~M. Rodriguez}, \bibinfo{author}{F.~Herrera}, \bibinfo{title}{2-tuple linguistic model}, \bibinfo{publisher}{Springer}, \bibinfo{year}{2015}.
\bibitem[{Zuheros et~al.(2023)}]{zuheros2023decision}
\bibinfo{author}{C.~Zuheros}, et~al.,
\newblock \bibinfo{title}{{Explainable Crowd Decision Making methodology guided by expert natural language opinions based on Sentiment Analysis with Attention-based Deep Learning and Subgroup Discovery}},
\newblock \bibinfo{journal}{Information Fusion} \bibinfo{volume}{97} (\bibinfo{year}{2023}) \bibinfo{pages}{101821}.
\bibitem[{Valdivia et~al.(2018)Valdivia, Luzón, Cambria, and Herrera}]{valdivia2018}
\bibinfo{author}{A.~Valdivia}, \bibinfo{author}{M.~V. Luzón}, \bibinfo{author}{E.~Cambria}, \bibinfo{author}{F.~Herrera},
\newblock \bibinfo{title}{Consensus vote models for detecting and filtering neutrality in sentiment analysis},
\newblock \bibinfo{journal}{Information Fusion} \bibinfo{volume}{44} (\bibinfo{year}{2018}) \bibinfo{pages}{126--135}.
\bibitem[{Huang et~al.(2023)Huang, Song, Wang, Chen, and Ma}]{huang2023look}
\bibinfo{author}{Y.~Huang}, \bibinfo{author}{J.~Song}, \bibinfo{author}{Z.~Wang}, \bibinfo{author}{H.~Chen}, \bibinfo{author}{L.~Ma},
\newblock \bibinfo{title}{Look before you leap: An exploratory study of uncertainty measurement for large language models},
\newblock \bibinfo{journal}{arXiv preprint arXiv:2307.10236}  (\bibinfo{year}{2023}).
\bibitem[{Arrieta et~al.(2020)}]{arrieta2020explainable}
\bibinfo{author}{A.~B. Arrieta}, et~al.,
\newblock \bibinfo{title}{{Explainable Artificial Intelligence (XAI): Concepts, taxonomies, opportunities and challenges toward responsible AI}},
\newblock \bibinfo{journal}{Information fusion} \bibinfo{volume}{58} (\bibinfo{year}{2020}) \bibinfo{pages}{82--115}.
\bibitem[{Bodria et~al.(2023)}]{bodria2023benchmarking}
\bibinfo{author}{F.~Bodria}, et~al.,
\newblock \bibinfo{title}{Benchmarking and survey of explanation methods for black box models},
\newblock \bibinfo{journal}{{Data Mining and Knowledge Discovery}}  (\bibinfo{year}{2023}) \bibinfo{pages}{1--60}.
\bibitem[{Longo et~al.(2024)}]{longo2024explainable}
\bibinfo{author}{L.~Longo}, et~al.,
\newblock \bibinfo{title}{{Explainable artificial intelligence (XAI) 2.0: A manifesto of open challenges and interdisciplinary research directions}},
\newblock \bibinfo{journal}{Information Fusion}  (\bibinfo{year}{2024}) \bibinfo{pages}{102301}.
\bibitem[{Kumar et~al.(2023)Kumar, Agarwal, Srinivas, Feizi, and Lakkaraju}]{kumar2023certifying}
\bibinfo{author}{A.~Kumar}, \bibinfo{author}{C.~Agarwal}, \bibinfo{author}{S.~Srinivas}, \bibinfo{author}{S.~Feizi}, \bibinfo{author}{H.~Lakkaraju},
\newblock \bibinfo{title}{{Certifying LLM safety against adversarial prompting}},
\newblock \bibinfo{journal}{arXiv preprint arXiv:2309.02705}  (\bibinfo{year}{2023}).
\bibitem[{Wei et~al.(2024)Wei, Haghtalab, and Steinhardt}]{wei2024jailbroken}
\bibinfo{author}{A.~Wei}, \bibinfo{author}{N.~Haghtalab}, \bibinfo{author}{J.~Steinhardt},
\newblock \bibinfo{title}{{Jailbroken: How does llm safety training fail?}},
\newblock \bibinfo{journal}{Advances in Neural Information Processing Systems} \bibinfo{volume}{36} (\bibinfo{year}{2024}).
\bibitem[{Ji et~al.(2024)Ji, Liu, Dai, Pan, Zhang, Bian, Chen, Sun, Wang, and Yang}]{ji2024beavertails}
\bibinfo{author}{J.~Ji}, \bibinfo{author}{M.~Liu}, \bibinfo{author}{J.~Dai}, \bibinfo{author}{X.~Pan}, \bibinfo{author}{C.~Zhang}, \bibinfo{author}{C.~Bian}, \bibinfo{author}{B.~Chen}, \bibinfo{author}{R.~Sun}, \bibinfo{author}{Y.~Wang}, \bibinfo{author}{Y.~Yang},
\newblock \bibinfo{title}{{Beavertails: Towards improved safety alignment of llm via a human-preference dataset}},
\newblock \bibinfo{journal}{Advances in Neural Information Processing Systems} \bibinfo{volume}{36} (\bibinfo{year}{2024}).

\end{thebibliography}
